\documentclass[twocolumn]{fairmeta}

\usepackage{bm}

\usepackage{lipsum}
\usepackage{colortbl}
\usepackage[percent]{overpic}
\usepackage{multirow}
\usepackage{array}
\usepackage{soul}
\newcommand{\PreserveBackslash}[1]{\let\temp=\\#1\let\\=\temp}
\newcolumntype{C}[1]{>{\PreserveBackslash\centering}p{#1}}
\newcolumntype{L}[1]{>{\PreserveBackslash\raggedright}p{#1}}

\usepackage{xcolor}

\newcommand{\ashutosh}[1]{{\color{teal}{}}}

\newif\ifshowtodo
\showtodotrue

\usepackage{tcolorbox}
\tcbuselibrary{breakable}

\definecolor{Gray}{gray}{0.90}

\newcommand{\OURS}{\textsc{HuDA}\xspace} %

\title{Human detectors are surprisingly powerful reward models}

\author[1,2,*]{Kumar Ashutosh}
\author[1]{XuDong Wang}
\author[1]{Xi Yin}
\author[2]{Kristen Grauman}
\author[1]{Adam Polyak}
\author[1]{Ishan Misra}
\author[1]{Rohit Girdhar}

\affiliation[1]{Meta}
\affiliation[2]{University of Texas at Austin}

\contribution[*]{Work done at Meta}

\abstract{

Video generation models have recently achieved impressive visual fidelity and temporal coherence. Yet, they continue to struggle with complex, non-rigid motions, especially when synthesizing humans performing dynamic actions such as sports, dance, \etc. Generated videos often exhibit missing or extra limbs, distorted poses, or physically implausible actions.
In this work, we propose a remarkably simple reward model, \OURS{}, to quantify and improve the human motion in generated videos.
\OURS{} integrates human detection confidence for appearance quality, and a temporal prompt alignment score to capture motion realism.
We show this simple reward function that leverages off-the-shelf models without any additional training, outperforms specialized models finetuned with manually annotated data.
Using \OURS{} for Group Reward Policy Optimization (GRPO) post-training of video models, we significantly enhance video generation, especially when generating complex human motions, outperforming state-of-the-art models like Wan 2.1, with win-rate of 73\%. Finally, we demonstrate that \OURS{} improves generation quality beyond just humans, for instance, significantly improving generation of animal videos and human-object interactions.

}

\date{\today}
\correspondence{Kumar Ashutosh at \email{kumar.ashutosh@utexas.edu}}

\metadata[Project Page]{\url{https://huda-reward-model.github.io/}}

\begin{document}

\maketitle
\section{Introduction}
\label{sec:intro}

\newlength\imggap
\setlength\imggap{5pt} %

\newlength\imgwSC
\setlength\imgwSC{\dimexpr(\columnwidth - 3\imggap)/4\relax}

\newcommand{\rowpad}{\vspace{0pt}}
\newcommand{\justcell}[1]{\parbox[t]{\dimexpr\imgwSC + 0.5\imggap\relax}{\raggedright\color{red}\scriptsize #1}}
\newcommand{\justcellblack}[1]{\parbox[t]{\dimexpr\imgwSC + 0.5\imggap\relax}{\raggedright\color{black}\scriptsize \emph{#1}}}

\begin{figure}[t]\footnotesize
\centering
\setlength{\tabcolsep}{0pt}
\renewcommand{\arraystretch}{1}

\begin{tabular}{@{\hspace{0.5\imggap}}C{\imgwSC}@{\hspace{0.5\imggap}}C{\imgwSC}@{\hspace{0.5\imggap}}C{\imgwSC}@{\hspace{0.5\imggap}}C{\imgwSC}@{\hspace{0.5\imggap}}}
\justcellblack{Dynamic flip} &
\justcellblack{~~Gymnastics} &
\justcellblack{~~~~Hopping} &
\justcellblack{~~Jump and roll}
\end{tabular}

\begin{tabular}{@{}c@{\hspace{\imggap}}c@{\hspace{\imggap}}c@{\hspace{\imggap}}c@{\hspace{\imggap}}c@{}}

\begin{overpic}[trim=100 270 50 270,clip,width=\imgwSC,keepaspectratio]{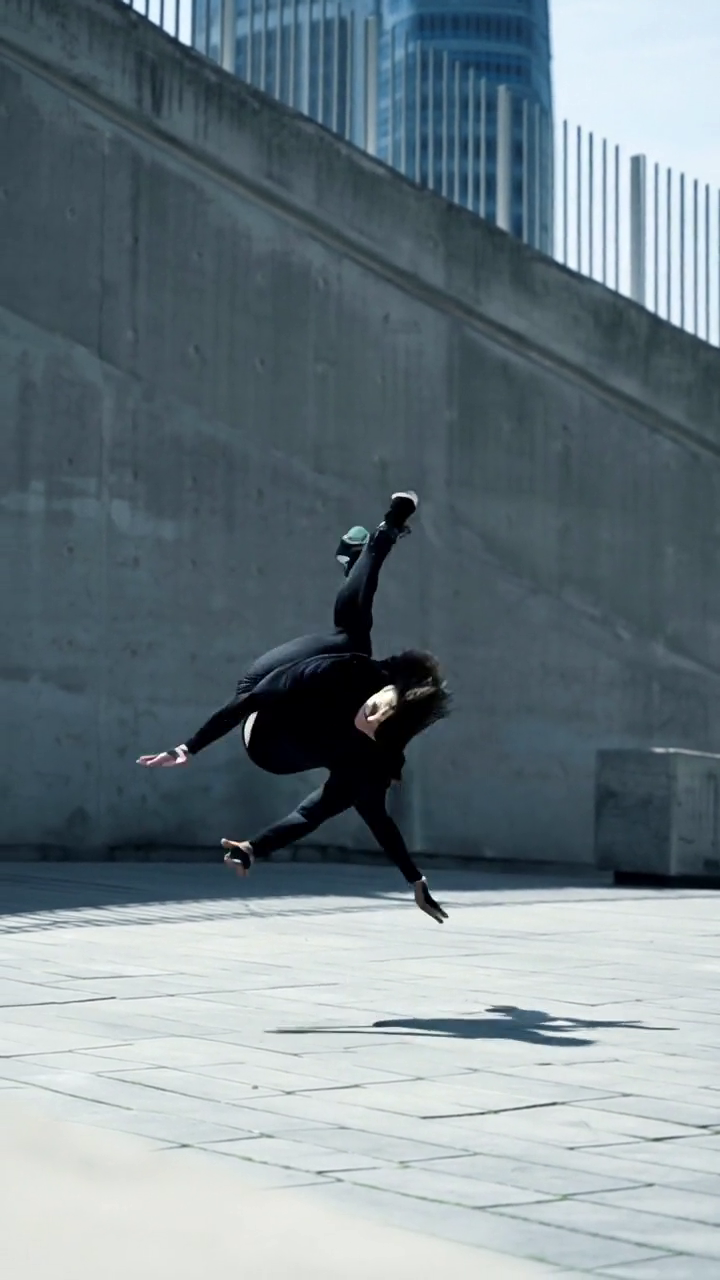}
  \put(3,90){\color{white}\scriptsize\bfseries Wan 2.1 14B}
\end{overpic} & 
\includegraphics[trim=0 170 200 430,clip,width=\imgwSC,keepaspectratio]{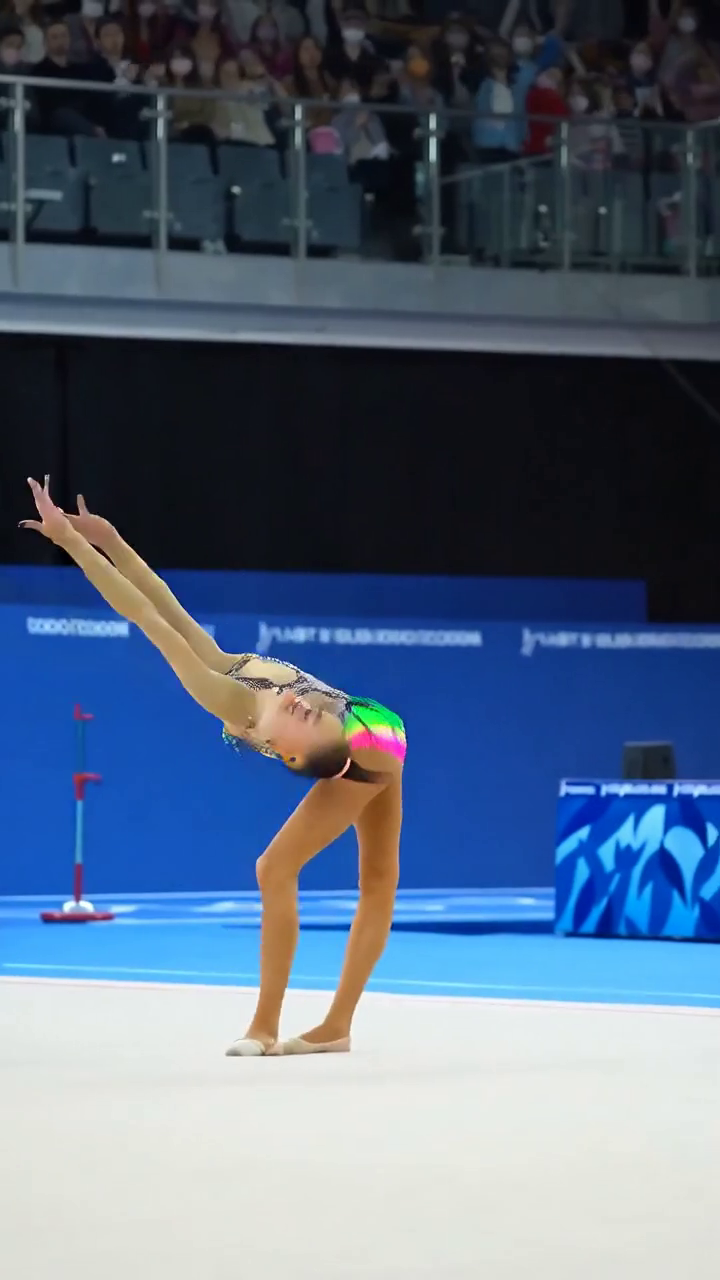} &
\includegraphics[trim=150 200 50 400,clip,width=\imgwSC,keepaspectratio]{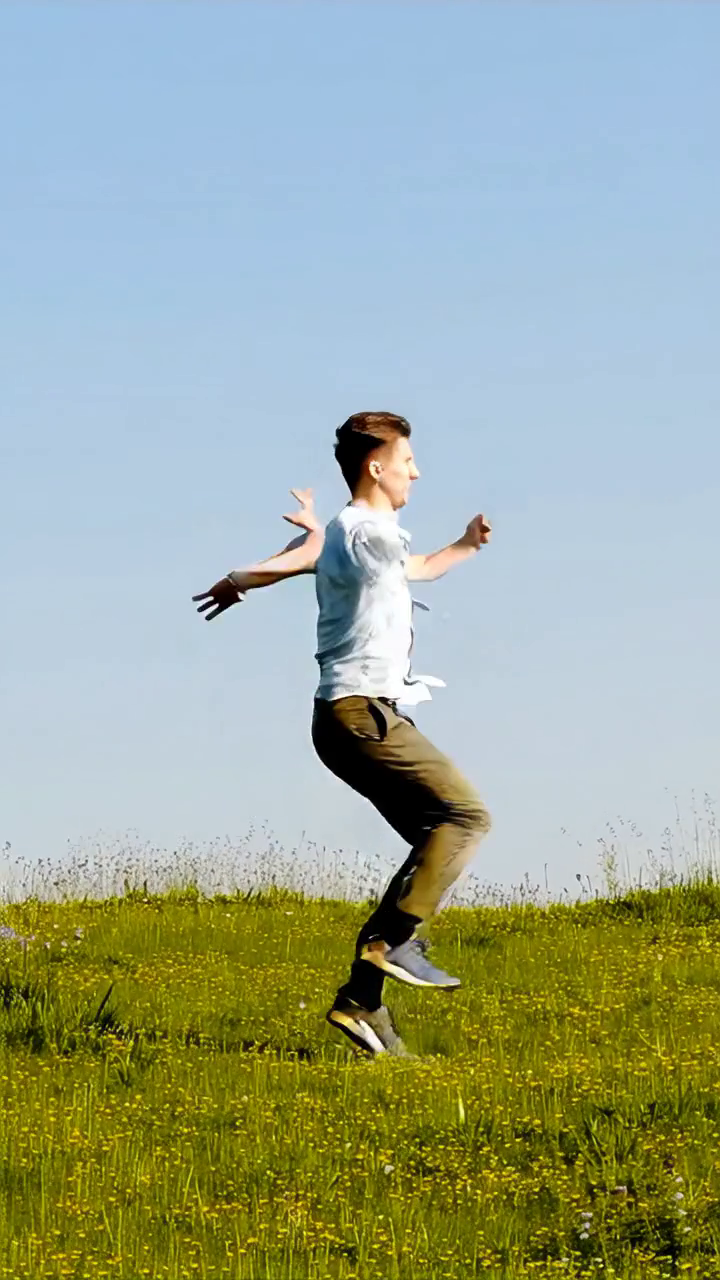} &
\includegraphics[trim=0 190 0 150,clip,width=\imgwSC,keepaspectratio]{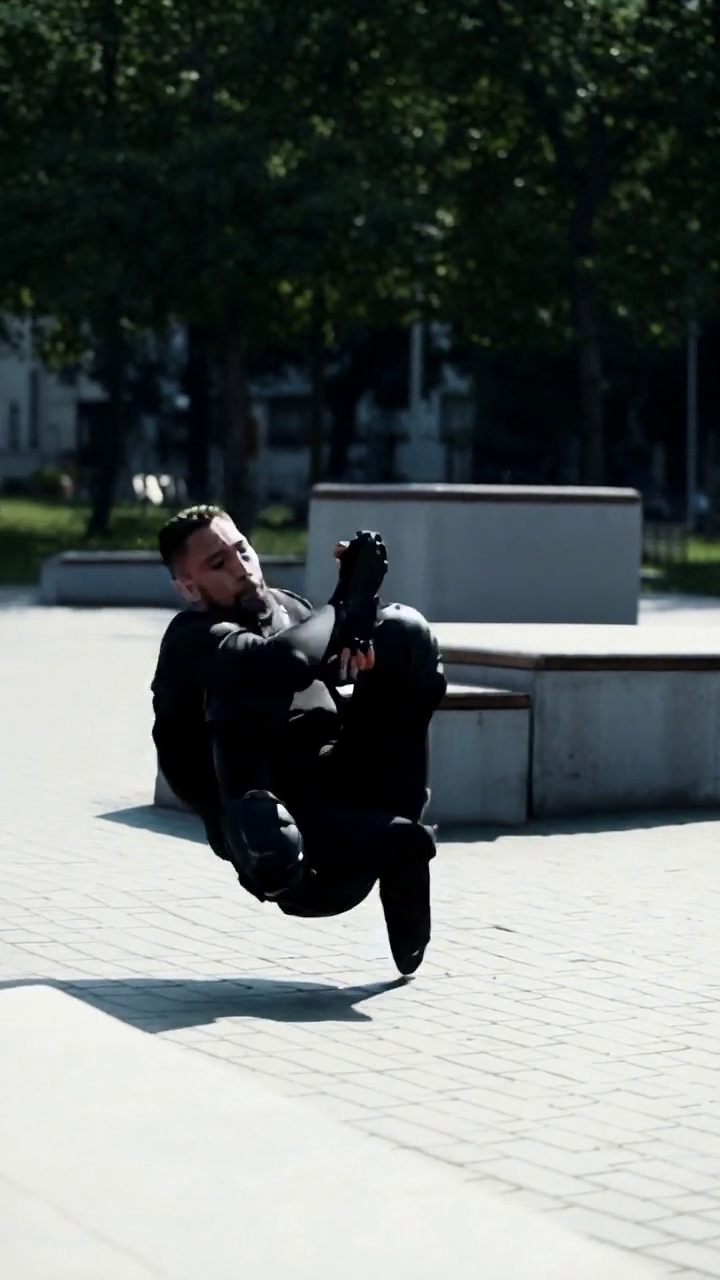}
\end{tabular}
\vspace{-10pt}

\begin{tabular}{@{\hspace{0.5\imggap}}C{\imgwSC}@{\hspace{0.5\imggap}}C{\imgwSC}@{\hspace{0.5\imggap}}C{\imgwSC}@{\hspace{0.5\imggap}}C{\imgwSC}@{\hspace{0.5\imggap}}}
\justcell{\tiny{Implausible pose}} &
\justcell{\tiny{Distorted body}} &
\justcell{\tiny{~Extra body part}} &
\justcell{\tiny{~Missing body part}}
\end{tabular}

\vspace{2pt}

\setlength{\heavyrulewidth}{0.1pt}   %
\arrayrulecolor{gray}  

\rowpad

\begin{tabular}{@{}c@{\hspace{\imggap}}c@{\hspace{\imggap}}c@{\hspace{\imggap}}c@{\hspace{\imggap}}c@{}}
    \midrule
\begin{overpic}[trim=100 220 50 310,clip,width=\imgwSC,keepaspectratio]{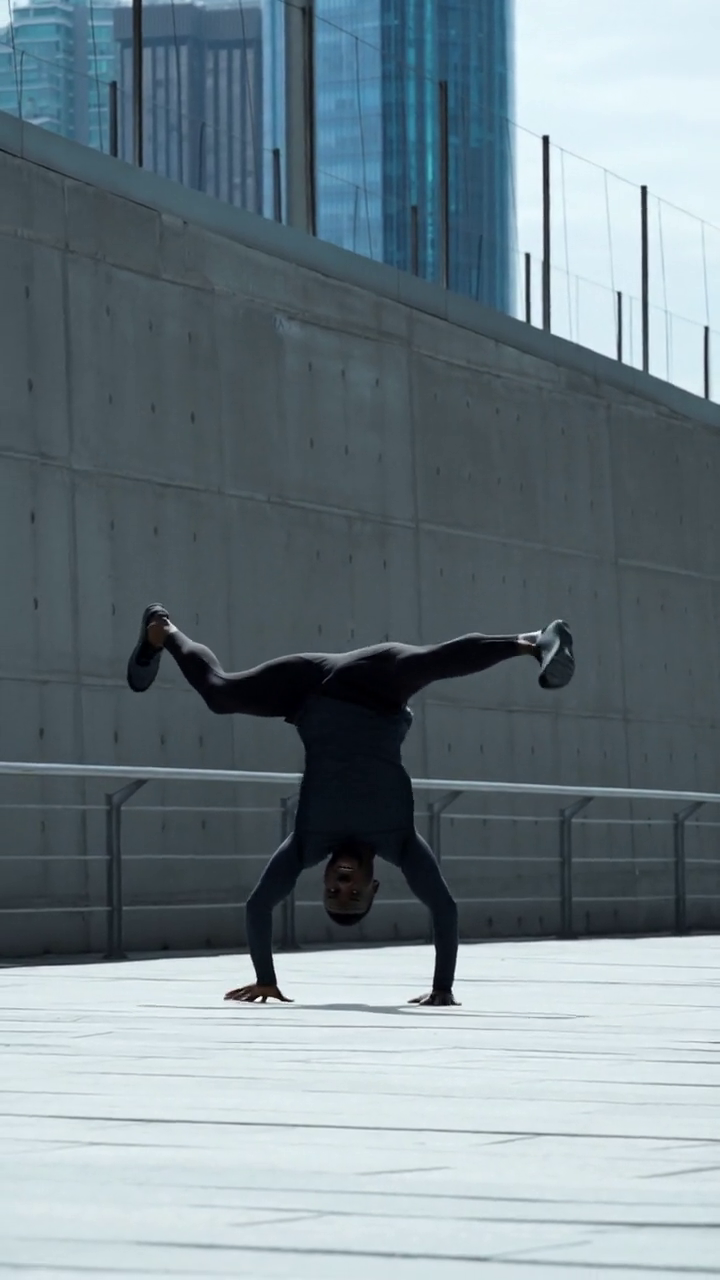}
  \put(3,90){\color{white}\scriptsize\bfseries \OURS (Ours)}
\end{overpic} & 
\includegraphics[trim=100 150 0 320,clip,width=\imgwSC,keepaspectratio]{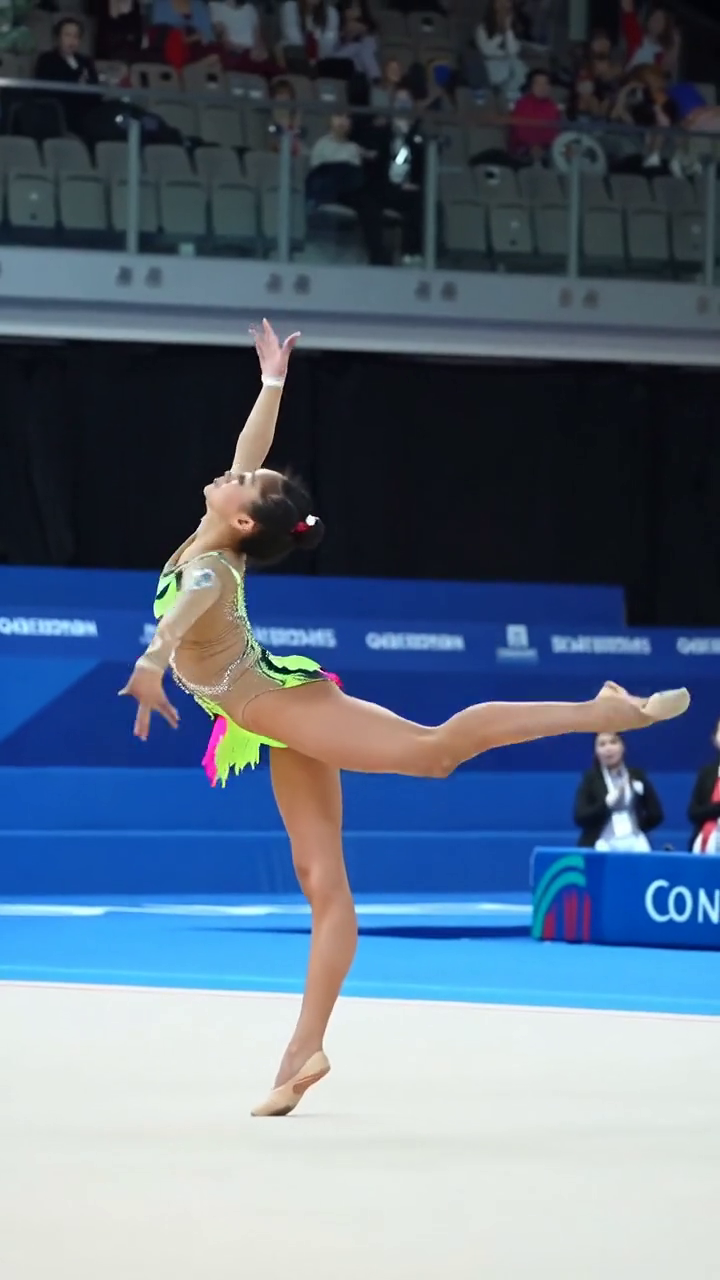} &
\includegraphics[trim=100 400 200 330,clip,width=\imgwSC,keepaspectratio]{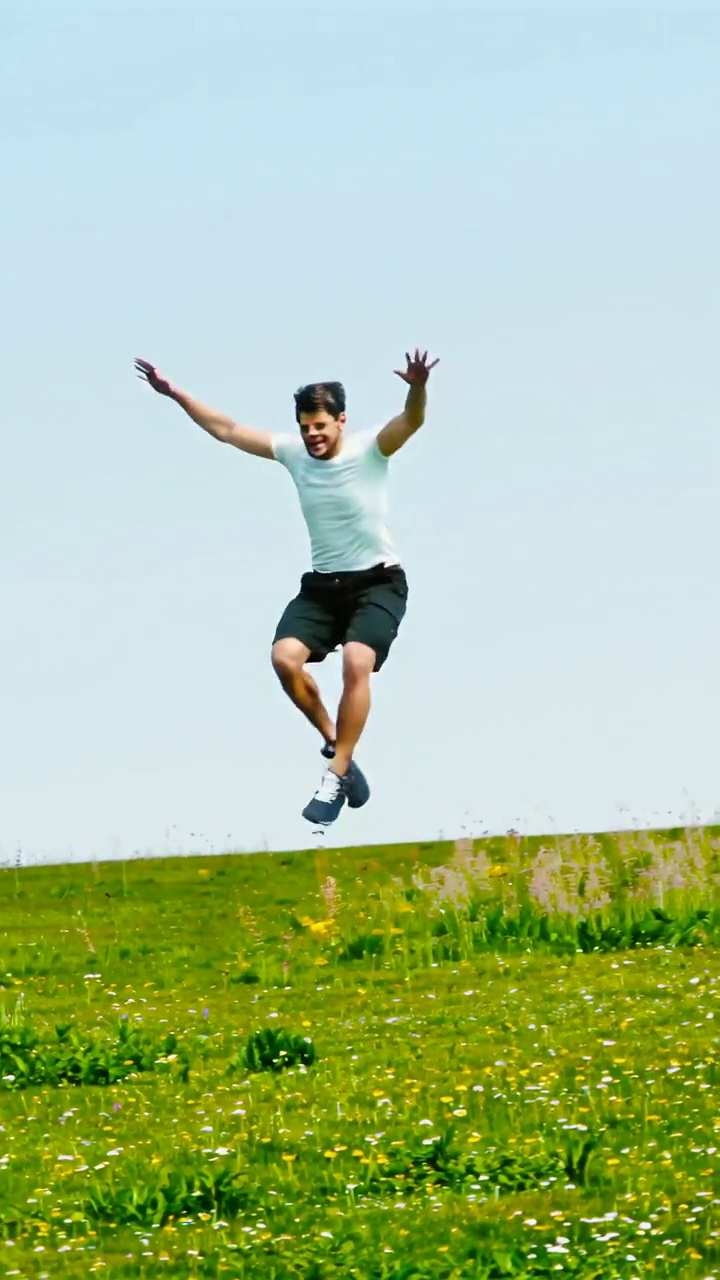} &
\includegraphics[trim=100 270 0 200,clip,width=\imgwSC,keepaspectratio]{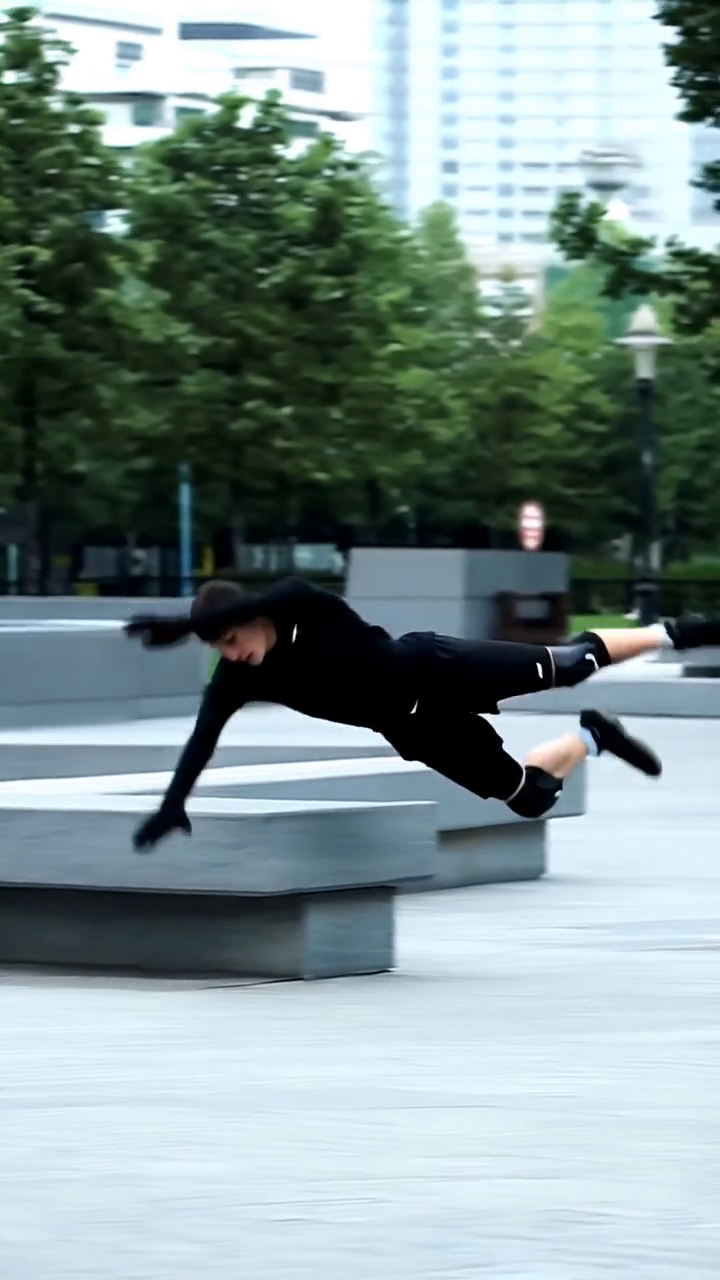}
\end{tabular}

\rowpad

\caption{{\bf \OURS} is a reward model to quantify the human appearance and motions in generated videos. \OURS detects extra or missing body parts, implausible body pose, and misalignment with the prompt. Training a video diffusion model with \OURS using GRPO results in improved human appearance and motion in generated videos (bottom) compared to state-of-the-art baseline (top). Videos corresponding to all figures in the paper are provided in the~\cref{sec:appdx:videos}.}
\label{fig:teaser}
\vspace{-0.5cm}
\end{figure}

\begin{figure}[t]
  \centering
  \setlength{\tabcolsep}{1pt}
  \begin{tabular}{cccccc} %
    \includegraphics[width=0.16\linewidth]{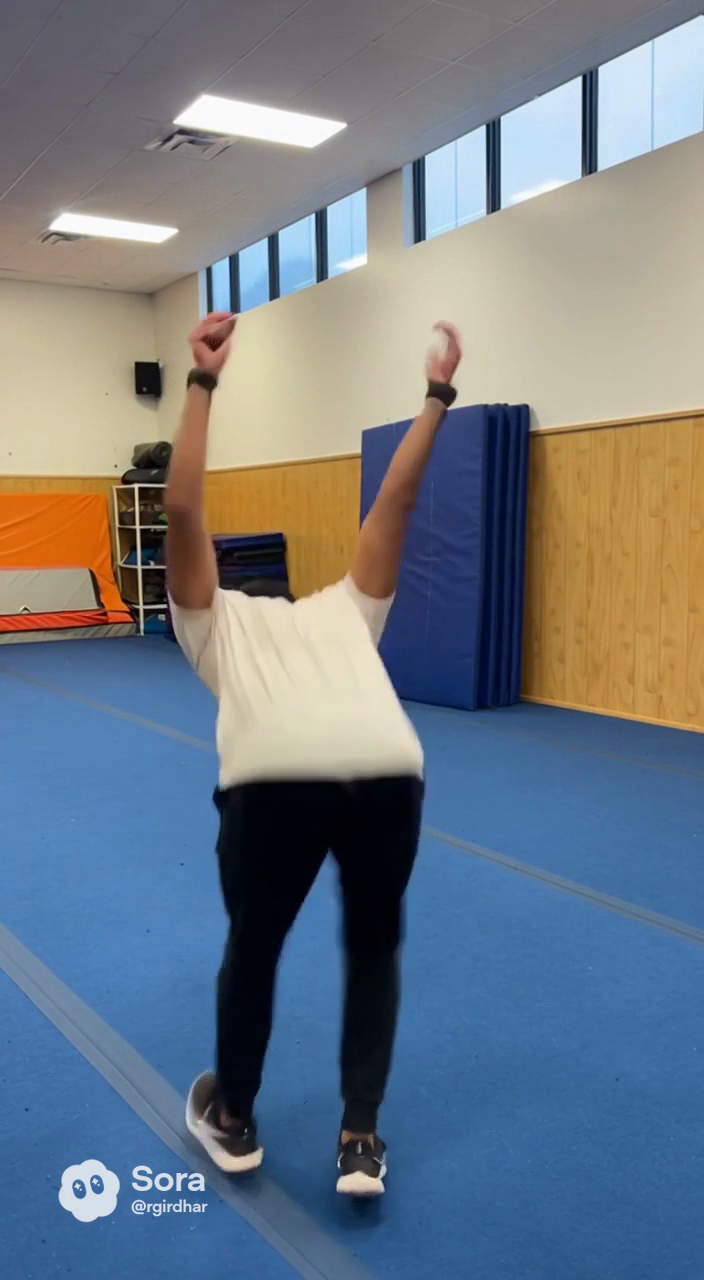} &
    \includegraphics[width=0.16\linewidth]{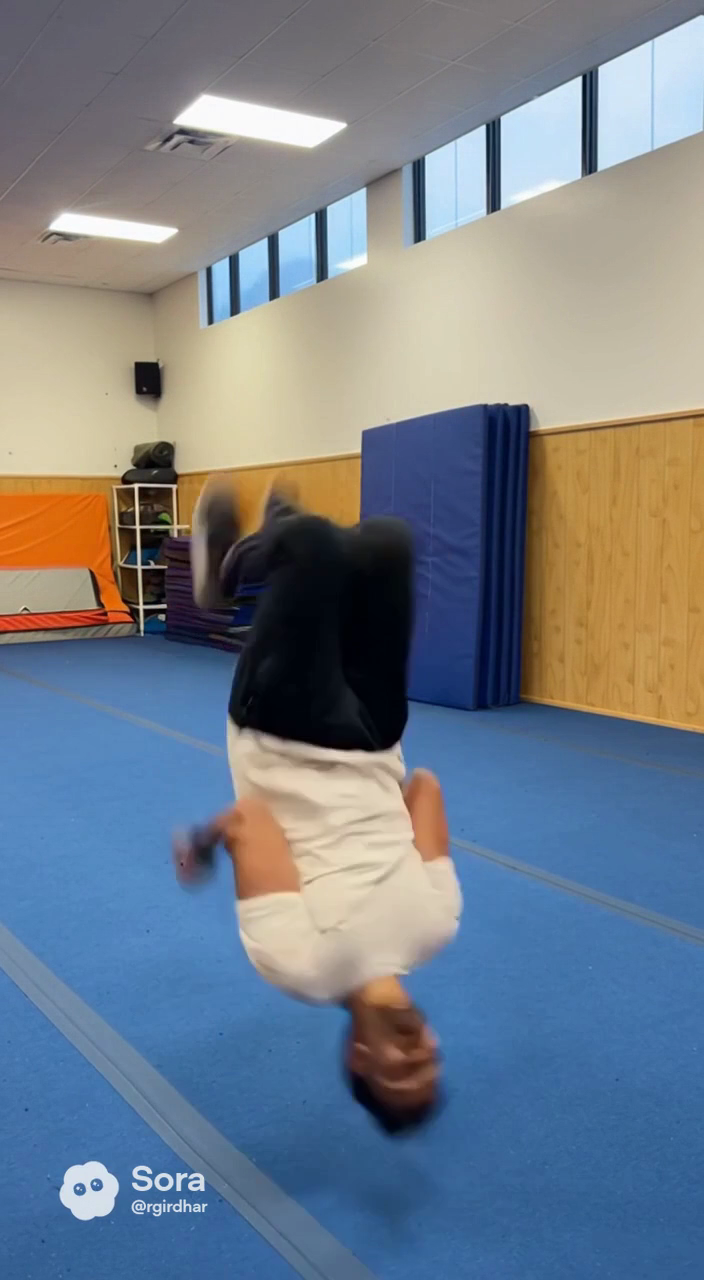} &
    \includegraphics[width=0.16\linewidth]{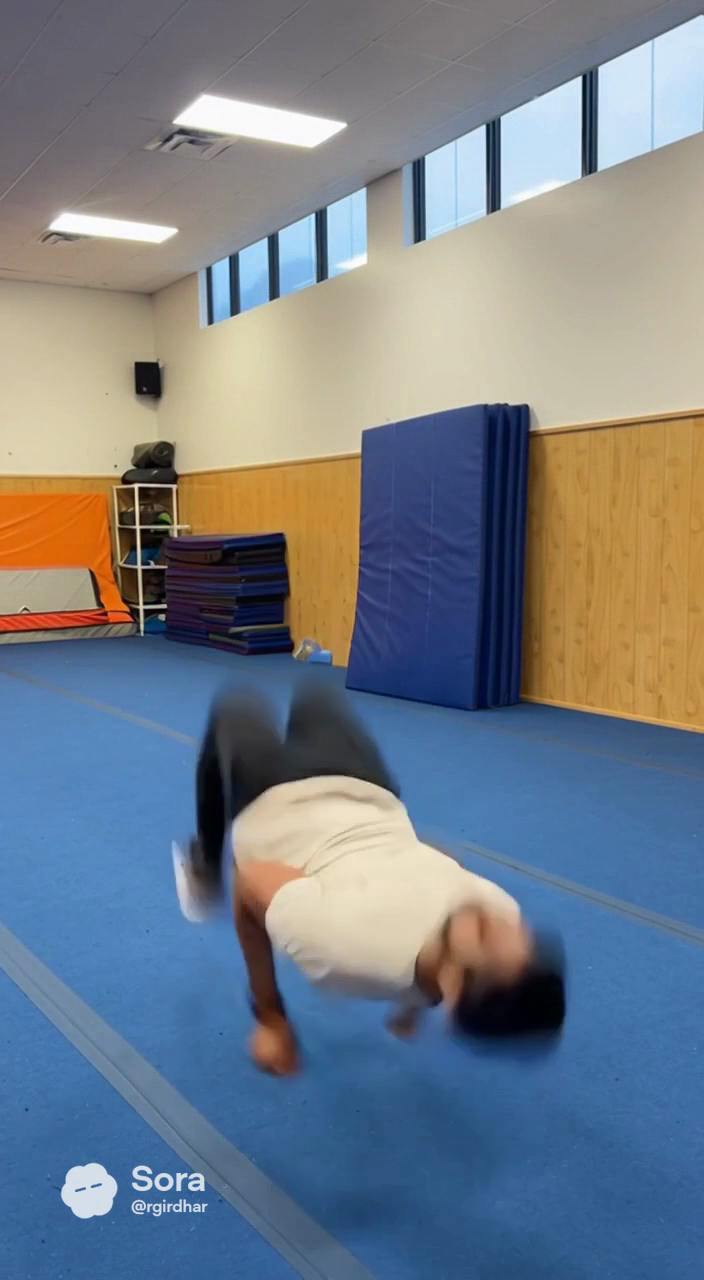} &
    \includegraphics[width=0.16\linewidth]{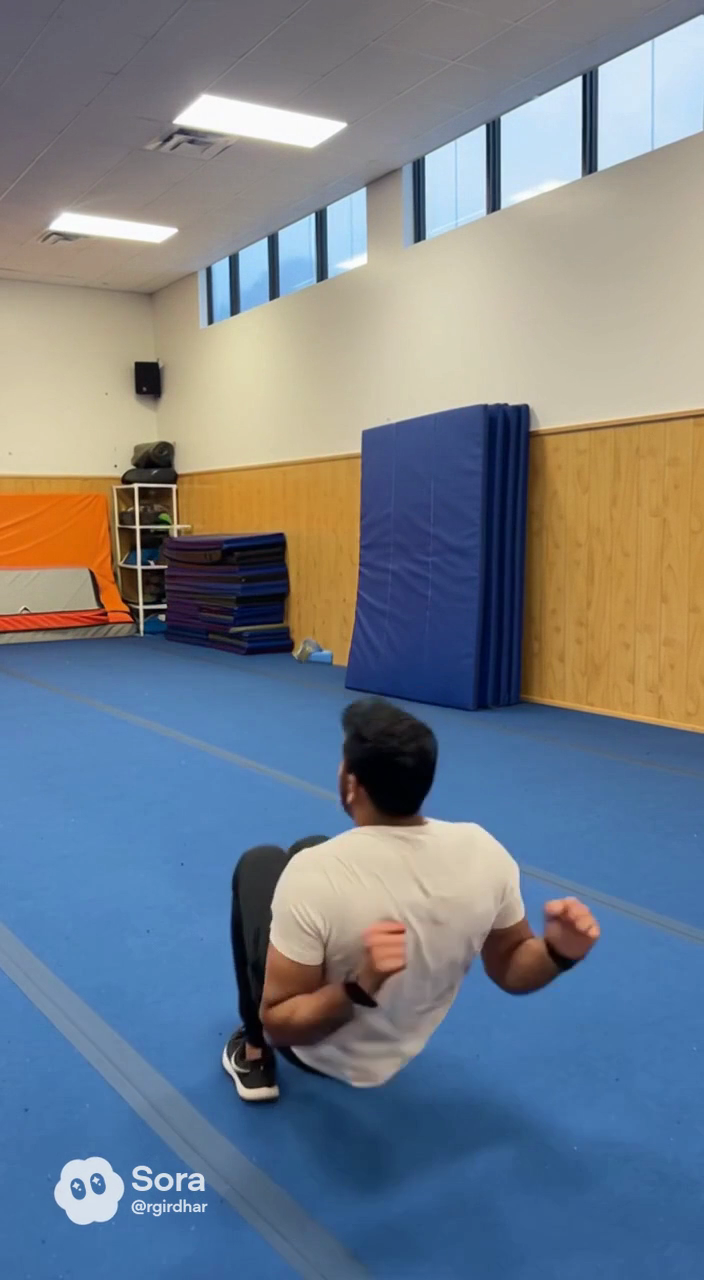} &
    \includegraphics[width=0.16\linewidth]{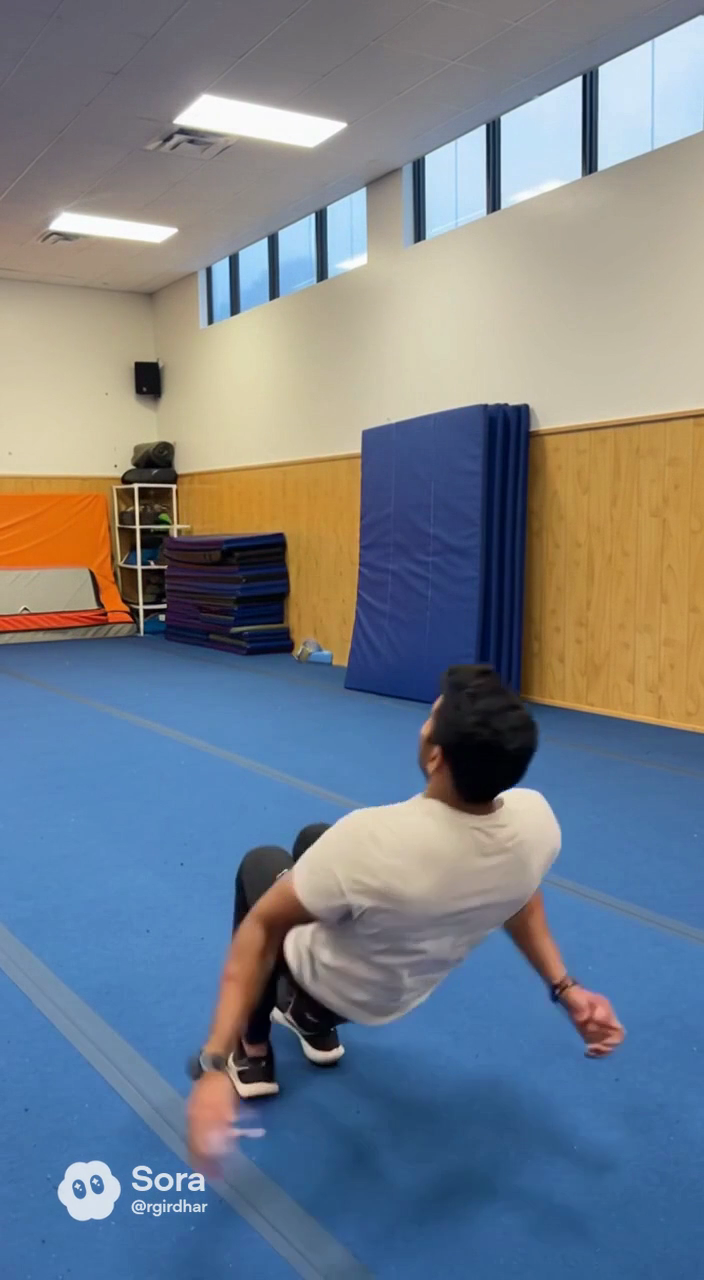} &
    \includegraphics[width=0.16\linewidth]{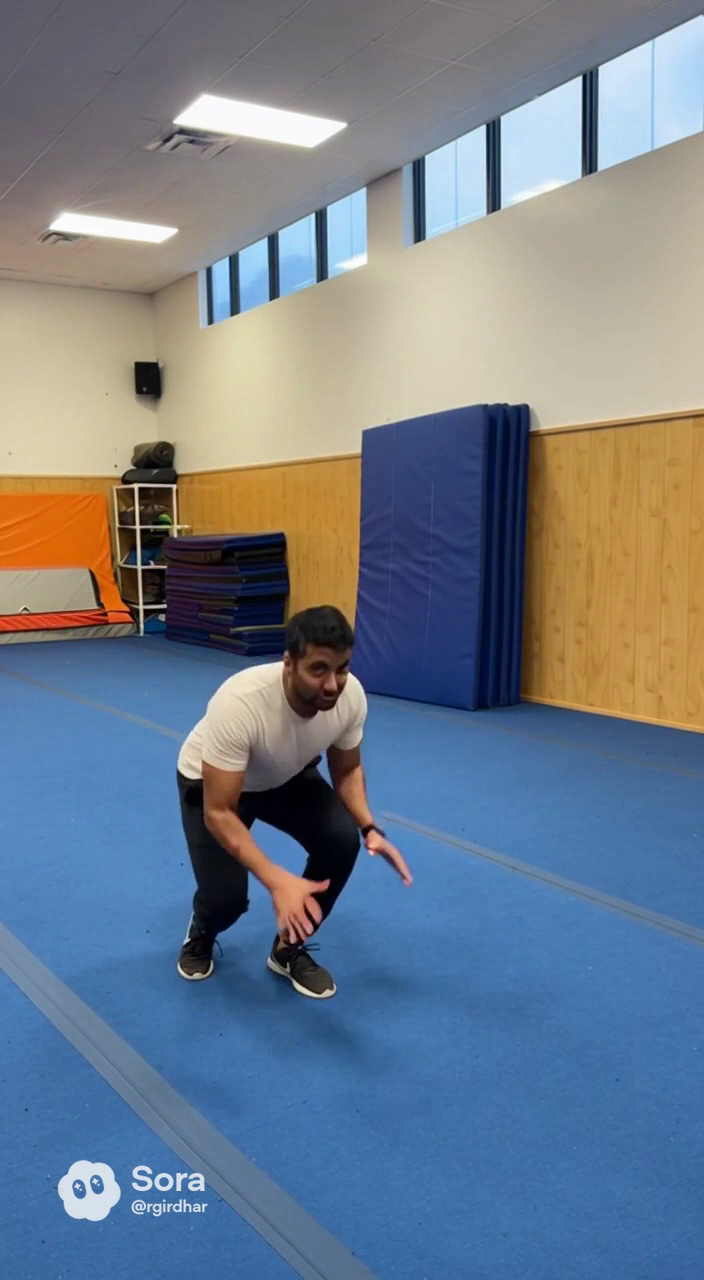} \\
    \includegraphics[width=0.16\linewidth]{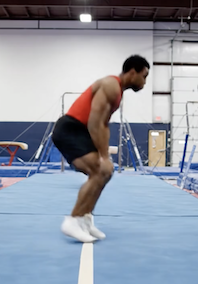} &
    \includegraphics[width=0.16\linewidth]{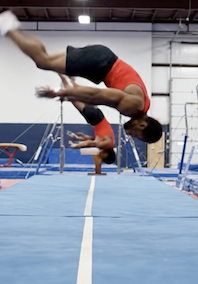} &
    \includegraphics[width=0.16\linewidth]{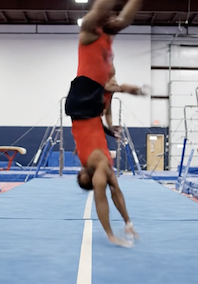} &
    \includegraphics[width=0.16\linewidth]{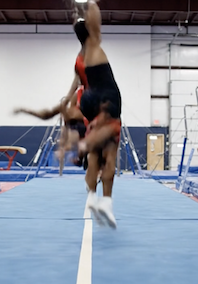} &
    \includegraphics[width=0.16\linewidth]{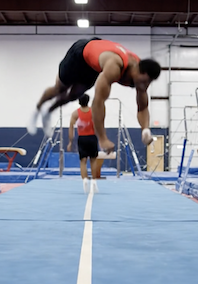} &
    \includegraphics[width=0.16\linewidth]{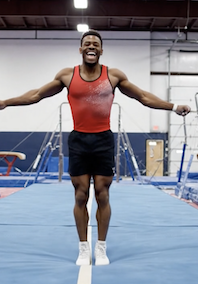} \\
  \end{tabular}
  \caption{{\bf SORA-2 and Veo-3 generations} for the prompts ``@rgirdhar (author) doing a backflip,'' and ``A person doing a backflip in a gym.'' Even state-of-the-art proprietary models struggle to generate complex human actions. Full video in~\cref{sec:appdx:videos}.
  }
  \label{fig:sora_altman_backflip}
  \vspace{-0.5cm}
\end{figure}

Text to video generation is an active area of research with growing applications across the industry, including in education, film-making, marketing, and social media.
From an educational app to help one master their gymnastic moves, to a fun social media trend showing one perform parkour, there is a particular emphasis on generating high fidelity videos of humans.
Despite impressive progress in the past few years, %
the generations do not always satisfy the desired requirements.
While recent work has developed measures to quantify and fix inaccuracies across some axes like prompt faithfulness or overall quality, it has been very challenging to generate accurate human shape and motion, especially when the human is performing complex actions (\cf~\cref{fig:teaser} (top)). %
This is true even for advanced proprietary models such as SORA-2 and VEO-3 (\cf~\cref{fig:sora_altman_backflip}).

In this work, we aim to address this challenging task of assessing and generating videos with complex human actions.
We propose a remarkably simple reward function to assess the quality of generations, {\bf \OURS{}}:
that leverages {\bf Hu}man {\bf D}etection and temporal prompt {\bf A}lignment. We find that the frame-level object detection~\cite{vitdet} confidence score for the `human' class is a strong indicator of how realistic the human in the generated video is. By itself, however, this reward is easy to hack---a low-motion video containing the ``simpler'' part of the action can obtain a perfect score by generating the human poses perfectly. For example, a video showing only the takeoff phase of the backflip, before the person flips in mid-air, the latter of which is typically harder for models to generate. To counteract this, we
strengthen the motion assessment by decomposing the input text prompt into ``phase captions'' corresponding to different phases of an action, and evaluating the similarity of these captions with uniformly sampled frames. Finally, we aggregate frame-level confidences for both models into a video-level reward score.

The critical challenge in designing a reward model to evaluate the realism of generated human actions is to understand and quantify all the failure cases, including extra or missing body parts, implausible body pose like knee bending forward, and implausible movements like a very high jump or a sliding effect without any force. Using a video language model (VLM)~\cite{qwen2.5} is not enough, as we also show in our experiments. There are prior works in identifying invalid human generations---missing or redundant body parts~\cite{generated_or_real}, or incorrect physics~\cite{finephys}. There are also efforts to solve this problem in a data-driven way by annotating a large amount of images if they have an anomaly~\cite{humanrefiner,heie,generated_or_real}, and training models on that.
\OURS, on the other hand, is a much simpler approach that does not require any training, and leverages off-the-shelf computer vision models in a zero-shot manner to build a powerful reward function.

Being able to evaluate the correctness of human motion and appearance also enables \emph{post-training} video diffusion models with \OURS as a reward model using reinforcement learning techniques such as Direct Preference Optimization (DPO)~\cite{diffusiondpo,videodpo} or Group Reward Policy Optimization (GRPO)~\cite{flowgrpo,dancegrpo}. This idea was also recently used to improve visual and motion quality, and prompt faithfulness of video models~\cite{videodpo,diffusiondpo,dancegrpo,flowgrpo}.

We use \OURS to tune video diffusion models using GRPO to improve complex human action generation. We experiment with state-of-the-art models such as Wan 2.1~\cite{wan} 1.3B and 14B variants, although our approach is applicable to any video diffusion model. For evaluation, we use a human preference study to first establish that our reward model strongly correlates with human judgement, compared to all the prior works. Next, we conduct another set of human preference studies to establish the improvements in the human appearance and motion in the generated videos using \OURS.
We show significant improvements over already strong models like Wan 2.1 14B, with win-rate of $73\%$, setting a new state-of-the-art in video generation for complex human actions.
Finally, we show that the training guides the model to better understand general bodily deformations, thus leading to improvements in animal video generations and human object interactions, without any specific training or optimization.

\section{Related Work}
\label{sec:related}

\textbf{Video generation.}
Research in video generation aims to generate a video, given a text prompt, or other conditioning signals like an image or a short video.
Traditional approaches use GAN~\cite{conditionalgan,imaginator}, pixel-level regression~\cite{jin2020exploring,geng2022comparing}, and other variational/stochastic models~\cite{denton2018stochastic,videoflow,han2022show}. Recent advances in video diffusion models~\cite{ltx,hunyuanvideo,wan,cogvideo,cogvideox,VDM,lvdm,imagen,make-a-video,align_latents,zstext2video,modelscope,latent_shift,animatediff} have enabled high-quality, realistic generations. 
Current methods are able to generate videos that are true to the conditioning signal, while still ensuring diversity with different starting input noise.

Despite the impressive progress, diffusion models are typically trained only using a simple L2 loss, and do not have additional visual, motion, human appearance objectives. Consequently, the outputs are not optimized on all crucial axes---text faithfulness, visual quality, and importantly, human appearance and actions. Prior works also observe these shortcomings of video diffusion models \cite{dancegrpo,flowgrpo,videodpo}. While most of these observations result in improvement along visual quality and prompt faithfulness, we design a novel training setup to train a text-to-video diffusion model to generate videos with correct human appearance without abnormalities.

\noindent \textbf{Generating videos of people.}
Prior works acknowledge issues with human appearance in images and videos \cite{humanrefiner,vbench2,finephys,generated_or_real,heie}. FHAD~\cite{generated_or_real}, HumanRefiner~\cite{humanrefiner}, and HEIE~\cite{heie} are manually annotated datasets containing real and generated images, with the label of whether the shown human is realistic or not. These datasets are then used to train models that can score the correctness of human appearance in generated images and videos~\cite{vbench2,generated_or_real,humanrefiner}. Alternatively, FinePhys~\cite{finephys} incorporates physics learned from gymnasium videos to improve in-domain video generation with more realistic gymnasium sequences. HumanDreamer~\cite{humandreamer} decouples the video generation problem to text-to-pose and pose-to-video generation, while manually annotating the pose representation for noise-free two-stage training. Finally, HEIE~\cite{heie} proposes a chain-of-thought (CoT) multimodal large language model (MLLM) to determine whether the generated image contains implausible human poses, relying on human annotated data.

In contrast to prior work, we propose a novel simple reward model that can score the plausibility of a human in the generated video. We do not make any assumption about the task the person is doing, unlike~\cite{finephys}, and is zero-shot, unlike~\cite{humanrefiner,vbench2,heie,generated_or_real,humandreamer}. We also do not focus solely on data-driven physical parameter estimation, as in \cite{finephys}, keeping our approach generalizable. We further show the effectiveness of the reward model by using it to post-train video diffusion models with GRPO to generate realistic human appearance and movement.

\noindent \textbf{Diffusion model post-training.} Training a diffusion model from scratch to optimize on certain axes is computationally expensive. Rather, recent methods use \emph{post-training} approaches to align the generated image or video outputs towards the desired behavior \cite{dancegrpo,flowgrpo,diffusiondpo,videodpo}. Several reinforcement learning (RL) approaches are used depending on the setup. Direct preference optimization (DPO) directly optimizes model outputs against preference pairs, eliminating the need for an explicit reward model and policy optimization stage \cite{diffusiondpo,na2025boost,curriculumdpo,videodpo}. Typically, the preference pairs are obtained via human annotations, but can also be obtained using a reward model \cite{videodpo}. On the other hand, group relative policy optimization (GRPO) uses an explicit reward model to update the diffusion model policy \cite{dancegrpo,flowgrpo}. In this work, we prefer GRPO over DPO for its on-policy training, the use of continuous richer reward signal, and the ability to use a multi-objective reward.

\section{Method}
\label{sec:method}

In this section, we first formally describe the problem statement, followed by our proposed approach.

\subsection{Problem definition}

Given videos $\bm{V}_1$ and $\bm{V}_2$, we aim to design a reward function,
$R$, such that $R(\bm{V}_1) > R(\bm{V}_2)$ iff $\bm{V}_1$ has better human appearance and motion, compared to $\bm{V}_2$, while portraying the same action. In particular, the characteristics of a good human appearance and motion include no extra or missing body part,
no abnormal body pose,
and  no abnormal body movement. %
In other words, any cue that allows a viewer to classify the video as implausible based on the human in the video should result in a lower value of the function $R$.
Our choice of $R$ is a superset of the considerations in prior work, which only focuses on missing and extra body parts~\cite{generated_or_real}, or incorrect physics~\cite{finephys}.
In addition to designing $R$, %
we aim to improve video generation performance by using $R$ in a reinforcement learning pipeline as a reward model. In particular, we show improvement in text-to-video (T2V) generations by using $R$ in GRPO. While our focus in this work is T2V, our approach is general and can be easily extended to other video generations tasks such as image-to-video (I2V) or video editing (V2V).

\begin{figure*}[t]\footnotesize
    \centering
    \includegraphics[width=\textwidth]{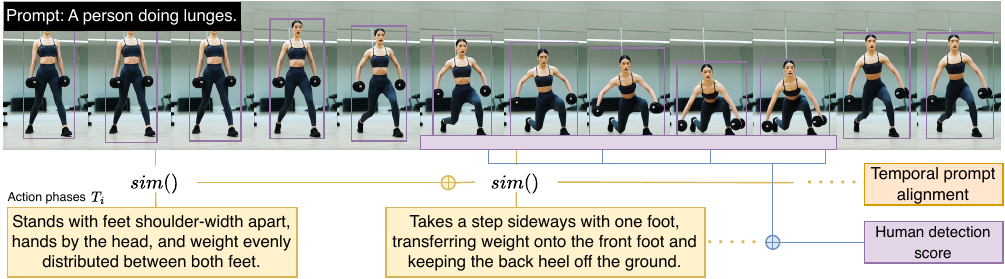}
    \caption{\textbf{Our reward model \OURS} quantifies the human detection score and temporal prompt alignment scores. The first term is the confidence level of the human detector in the worst performing window $W$ shown in purple band. 
    Next, temporal prompt alignment score is the average of similarity between phase descriptions $T_i$ and uniformly sampled frames. Overall reward is a weighted average of human detection score and temporal prompt alignment.}
    \label{fig:reward_overview}
    \vspace{-0.5cm}
\end{figure*}

\subsection{Proposed reward model $R$}
We propose \OURS: a simple linear combination of a) {\bf Hu}man {\bf D}etection score, and b) temporal prompt {\bf A}lignment.
As we show in experiments later, this simple function combining off-the-shelf models captures human motion and appearance quality better than even approaches trained for this task. We now describe the components of this reward function, followed by how we use it for improving video generation in RL post-training.

\noindent \textbf{Human detection score} (H-score) measures the quality of humans in all the frames, and it is low if there are missing or extra body parts, implausible body pose, and incorrect physics. We measure the implausibility per frame, and then aggregate the scores into a video-level appearance quality.

Given an object detector model $\mathcal{D}: \mathbb{R}^{H \times W \times 3} \rightarrow O$  (\eg, ~\cite{maskrcnn,fastrcnn,vitdet}), where $O$ is the object name or the corresponding class, we quantify the detection score of $\bm{V}$
as

$$ \min_{\forall t} \left[ \sum_{\tau=t}^{t+W} \max_{\forall B} \left\{ \mathbb{P} \left( \mathcal{D}(\bm{V}_{\tau}[B]) = \text{human} \right) \right\} \right], $$
where $W$ is the window size, $B$ is any bounding box in the frame $\bm{V}_t$, and $\mathbb{P}$ refers to the probability of that detection. Concisely, the human detection score is the minimum confidence score over all fixed-size windows, where a window's confidence score is the sum of highest likelihood of the object detector predicting a human in the frame, for all frames in that window. We use minimum over average/max aggregation to capture the worst human appearance in a video. Note that this scorer is zero-shot, \ie, $\mathcal{D}$ is never trained with any generated image or video, and thus, requires no manual annotation effort, unlike prior work~\cite{humanrefiner,vbench2,finephys,generated_or_real,heie}.

While human detection score is a strong measure, as we show in the experiments, using it as the only reward model will motivate the video generation model to suppress motion, since a static video with good human appearance will give a perfect score, despite not showing the overall motion required in the input text prompt. To this end, we propose an additional action progress score, defined next.

\noindent %
\textbf{Temporal prompt alignment} (P-score)
measures the \emph{faithfulness} of the video frames for various action phases. We extract the detailed action phases from the text prompt. For example, the action phases for a lunge can be standing straight with legs apart, bending, holding, and coming back up (\cf~\cref{fig:reward_overview}). Now, given $T_i$ as the text describing the $i^{th}$ phase of the condition, out of a total of $N$ phases, the alignment is defined as

$$ \sum_{i=1}^{N} \text{sim}(\bm{V}_{t^i}, T_i) \quad \text{where}\quad t^i = \frac{i*|\bm{V}|}{N} $$
where $\text{sim}()$ is a similarity measure of visual and text features~\cite{blip}. This alignment measures action progress by comparing the current frame with the corresponding condition $T_i$. We represent $T_i$ with text for its interpretability and the ease of generation using off-the-shelf LLMs~\cite{qwen,llama3}.

The overall reward model \OURS is the sum of human detection score and a weighing factor ($\alpha$) times the temporal prompt alignment. \cref{fig:reward_overview} shows the reward model calculation for a representative example.

\subsection{RL post-training with \OURS}

We use the Group Reward Policy Optimization (GRPO)~\cite{dancegrpo,deepseekmath} to post-train the video diffusion model to generate outputs with better human appearance. As outlined previously in the~\cref{sec:related}, we choose GRPO over DPO for its on-policy training, use of continuous reward, and the ability to use multi-objective reward.

In particular, the objective function for GRPO is
\begin{multline*}
\mathcal{J}(\theta) =
\mathbb{E}_{\bm{V}_i \sim \pi_{\theta_{\text{old}}}}
\left[
\frac{1}{G} \sum_{i=1}^{G}
\min\left(
    \mathcal{L}_u, \mathcal{L}_c
\right)
\right] \\
\text{where}~~\mathcal{L}_u = \rho_i A_i,~~~ \mathcal{L}_c  = \operatorname{clip}(\rho_i, 1 - \epsilon, 1 + \epsilon) A_i \\
\text{and}~\rho_i =
\frac{\pi_\theta(\bm{V}_i \mid \mathbf{c})}
{\pi_{\theta_{\text{old}}}(\bm{V}_i \mid \mathbf{c})},
 \quad
A_i =
\frac{R(\bm{V}_i) - \mu_R}{\sigma_R}
\end{multline*}
Here, $\pi_{\theta_{old}}$ and $\pi_{\theta}$ are the policy variants before and after the update, respectively, and $\bm{V}_i$ and $\mathbf{c}$ are the video and the conditioning signal, \ie, text prompt. $\mu_R$ and $\sigma_R$ are the mean and standard deviation of the rewards of the $G$ generated samples. $\mathcal{L}_u$ and $\mathcal{L}_c$ are the unclipped and the clipped losses, respectively. Conventional GRPO also has a KL-loss between the original policy and the current policy~\cite{deepseekmath,flowgrpo}. However, following \cite{dancegrpo}, we do not use KL-loss since there is no performance gain. This choice also helps save memory since we don't need to keep a copy of the original policy.

\subsection{Implementation details}

We use ViTDet~\cite{vitdet,detectron2} as the object detector~$\mathcal{D}$. The window size $W$ is $6$ frames. We choose $N=5$ phases, and generate progress descriptions using Llama-3.2 8B~\cite{llama3}. The prompt describes the action as spanning $5-7$ seconds, and asks for $N$ phase caption, each spanning $1-2$ seconds. The exact prompt is given in the~\cref{sec:appdx:llm_prompts}.
We use BLIP~\cite{blip} as the similarity measure $sim()$. Finally, we set the weighing parameter $\alpha=0.5$ to focus more on human detection score over temporal prompt alignment score.

For video generation training, we experiment with 1.3B and 14B variants of Wan 2.1~\cite{wan} for our experiments; the method can be extended to any video diffusion model~\cite{hunyuanvideo,mochi,ltx}.
We train the model using $32$ A100 GPUs for $12$ hours. The learning rate for 1.3B and 14B models are $10^{-5}$, and $10^{-6}$, respectively.
During GRPO training, we generate videos with $\text{[H, W, T]}~=~[480, 480, 53]$. We generate $24$ videos per prompt, and sample $16$ steps using SDE sampler~\cite{dancegrpo,deepseekmath}. During inference, we generate videos with $\text{[H, W, T]}~=~[1280, 720, 81]$. We use $50$-step sampling 
with classifier free guidance (CFG) scale $5.0$ for both training and testing. 
These inference parameter choices reflect the best possible setting for the backbone video model~\cite{wan}.

\section{Experiments}
\label{sec:results}

In this section, we first establish \OURS as the accurate reward model for human appearance and pose. Next, we demonstrate that using \OURS as a reward model in GRPO training improves human appearance in generated videos.
To that end, we first discuss the common experimental setup, followed by results. We also discuss ablations, and finally, generalizability of \OURS to improve human-object interactions, and even improved animal video generation.

\subsection{Experimental Setup}
\label{sec:expt-setup}

In this section, we discuss experimental setup that is common for both evaluating \OURS as a reward model, and verifying the improvement in video generation.

\noindent \textbf{Creating the training and testing prompt set.} We care about human appearance and action, and thus, we create a prompt set to train and test the same.
We start with a wide range of activities, representing diverse motions with varying difficulty. Specifically, we leverage a large language model, Llama-3.2-8B~\cite{llama3}, to generate $3000$ human motion sequences labels. These are categorized into three levels of motion difficulty---easy: standing, walking, clapping; medium: running, jumping, tennis forehand; and hard: backflip, handstand, ski flip. The exact input prompt to generate the motions is given in the~\cref{sec:appdx:llm_prompts} that describes the diverse range of motion requirements, with examples of easy, medium and difficult motions.

Once we obtain the labels, we use a simple template to convert the labels into full prompts. We use kid, boy, girl, man, and woman as the subject and curate prompts of the format \emph{``A \{subject\} doing \{action\}''}. Finally, we use the prompt expansion in Wan 2.1~\cite{wan} to add details to the prompt. We choose a subset of $300$ human motion sequences (out of $3000$) for evaluation. The evaluation set contains $100$ human motions from each difficulty level. The same evaluation split is used for both evaluating \OURS's performance as a reward model, and showing improved video generation with \OURS.

\begin{table}[t]\footnotesize
\centering
\begin{tabular}{l c}
\hline
\textbf{Method} & \textbf{Accuracy} \\
\hline
VLM-as-a-judge~\cite{qwen2.5}\quad & 55.0 \\
DanceGRPO~\cite{dancegrpo} & 51.5 \\
\midrule
VBench-2.0~\cite{vbench2} & 72.7 \\
\midrule
\OURS (Ours) & \textbf{77.4} \\
\hline
\end{tabular}
\caption{\textbf{Accuracy} of the various reward models in predicting human-preferred videos. First two rows are zero-shot, and VBench-2.0 is trained for this task. Ours is fully zero-shot and significantly outperforms both zero-shot and trained baselines.}
\label{tab:winrate-expt1}
\vspace{-0.5cm}
\end{table}

\noindent \textbf{Baselines.} We compare \OURS with the following:

\begin{itemize}
    \item \textbf{VLM-as-a-judge~\cite{qwen2.5}}: Video language models (VLM) are capable of a variety of tasks, showcasing impressive zero-shot performance. In this baseline, we evaluate if zero-shot VLMs can detect abnormal human appearance and motion. Given a video with human, we query the VLM if the video contains any appearance anomaly, prompting it to give yes/no answer. We use the `no' logit probability as the reward score. The exact prompt is given in the~\cref{sec:appdx:llm_prompts}.
    \item \textbf{VBench-2.0 human anomaly~\cite{vbench2}}: VBench-2.0 proposes various video generation benchmarks. In particular, human anomaly detector predicts the likelihood of the video containing anomalous human appearance or motion. The proposed model is trained from samples from HumanRefiner~\cite{humanrefiner} dataset, along with additional supervision obtained from real and generated videos.
    \item \textbf{Video visual/motion alignment~\cite{dancegrpo}}: Many recent works use VLM-based visual and motion quality scores to improve video. We use DanceGRPO~\cite{dancegrpo} as a representative example to investigate if a combination of visual and motion quality can quantify human appearance.
    \item \textbf{Base model (Wan 2.1)~\cite{wan}}: This original model is used for evaluating improvements in video generation.

\end{itemize}

\noindent In addition to proposing \OURS, we also explore if other modalities, \eg, 3D/2D pose~\cite{vitpose}, or optical flow~\cite{dpflow} can be used as reward functions. We discuss our early explorations in the~\cref{sec:appdx:addl_qual}.

\begin{figure}[t]\footnotesize
    \centering
    \includegraphics[width=\columnwidth]{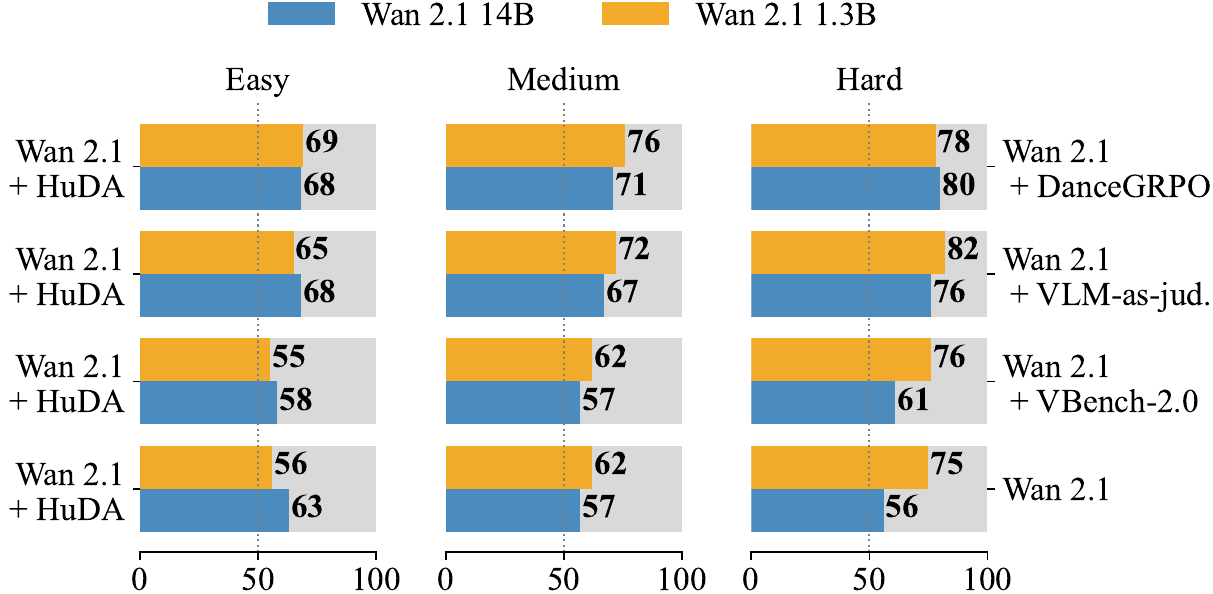}
    \caption{\textbf{Video generation win-rate}
    between \OURS (depicted by colored bars) and previous state-of-the-art baselines ($100 -$ \OURS's win rate). Our method is consistently preferred by human annotators across all settings, with biggest gains in the hard category (avg win-rate $73\%$ vs $63\%$ for easy, across all baselines).
    }
    \label{fig:qual}
    \vspace{-0.5cm}
\end{figure}

\noindent \textbf{Evaluation setup.} Our objective is to experimentally validate \OURS as the correct reward model, and then use \OURS to improve video generation.

Firstly, to evaluate \OURS as a reward model, we collect ground truth annotations from human raters.
In accordance to our task, we choose two videos for the same input prompt and ask the annotators to choose the video that has better human appearance. The prompts are from our test set, discussed above.
The instructions and examples demonstrate less preferred videos with extra/missing body parts, distorted or implausible poses.
We insert test examples to check for the attentiveness of the annotators. We have an annotator quality threshold to ensure correct alignment with human preference. Each sample is annotated by five annotators, and we choose the samples that have four annotators agreeing on an option. A screenshot of the annotation interface is given in the~\cref{sec:appdx:human_eval}.

\newlength\groupgap
\setlength\groupgap{8pt} %

\newlength\imgw
\setlength\imgw{\dimexpr(\textwidth-2\groupgap)/9\relax}

\newcommand{\rowcaption}[1]{%
  \noindent\par\vspace{3pt}%
  {\small #1}\par\vspace{6pt}%
}

\begin{figure*}[t]
\centering
\setlength{\tabcolsep}{0pt}
\renewcommand{\arraystretch}{1}

\begin{tabular}{@{}*{3}{c}@{\hspace{\groupgap}}*{3}{c}@{\hspace{\groupgap}}*{3}{c}@{}}
\multicolumn{3}{@{}p{\dimexpr3\imgw\relax}@{}}{\centering\footnotesize \emph{A parkour athlete performing a dynamic roll after landing from a jump in a city park.}} & 
\multicolumn{3}{@{}p{\dimexpr3\imgw\relax}@{}}{\centering\footnotesize \emph{A martial artist performing a dynamic 540-degree spinning kick in mid-air.}} & 
\multicolumn{3}{@{}p{\dimexpr3\imgw\relax}@{}}{\centering\footnotesize \emph{A gymnast performing a one-arm handstand spin on a gym floor.}} \\

\begin{overpic}[width=\imgw,keepaspectratio]{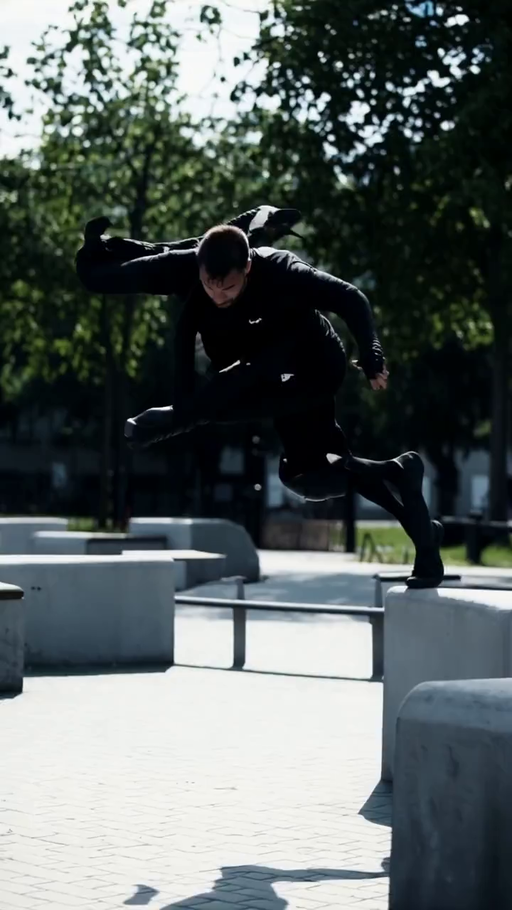}
  \put(3,7){\color{black}\tiny\bfseries Wan 2.1 14B}
\end{overpic} & 
\includegraphics[width=\imgw,keepaspectratio]{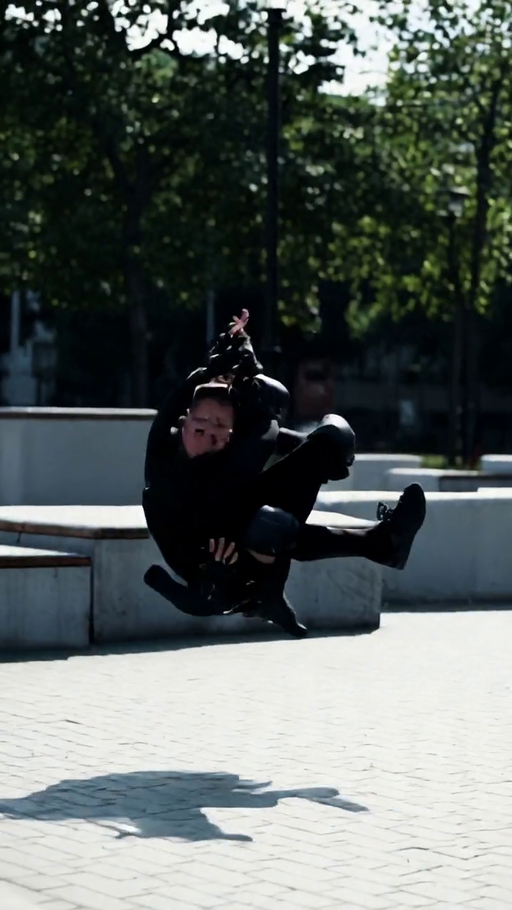} &
\includegraphics[width=\imgw,keepaspectratio]{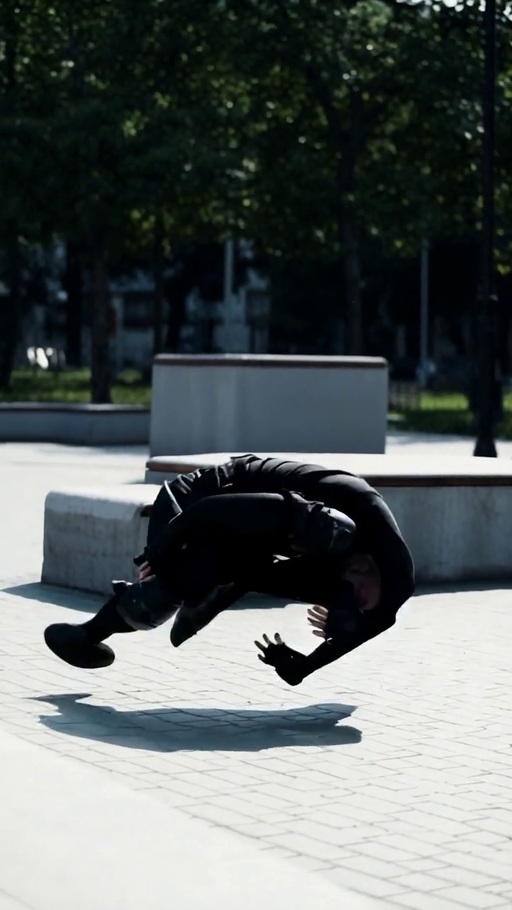} &
\begin{overpic}[width=\imgw,keepaspectratio]{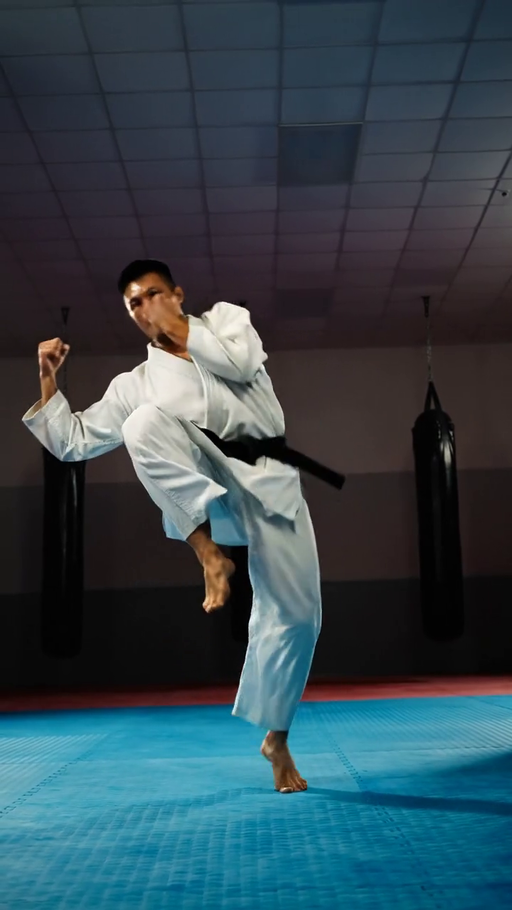}
  \put(3,5){\color{white}\tiny\bfseries Wan 2.1 14B}
\end{overpic} & 
\includegraphics[width=\imgw,keepaspectratio]{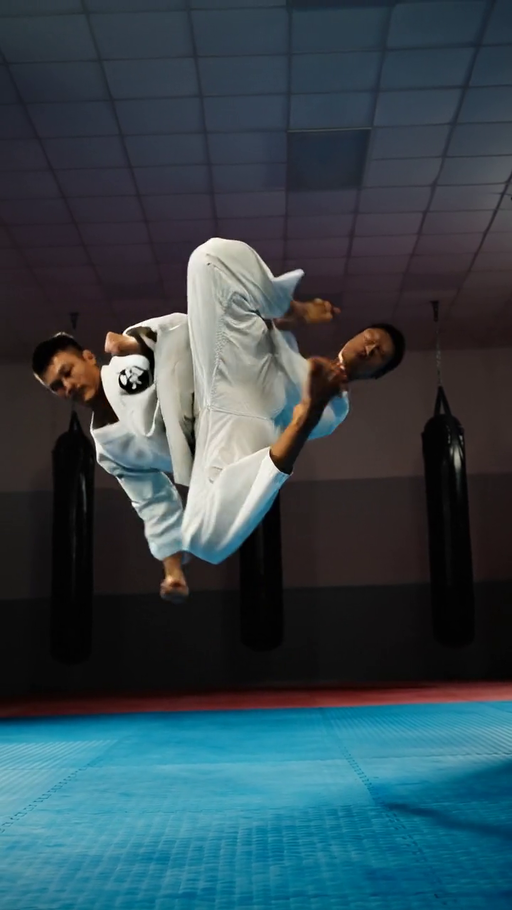} &
\includegraphics[width=\imgw,keepaspectratio]{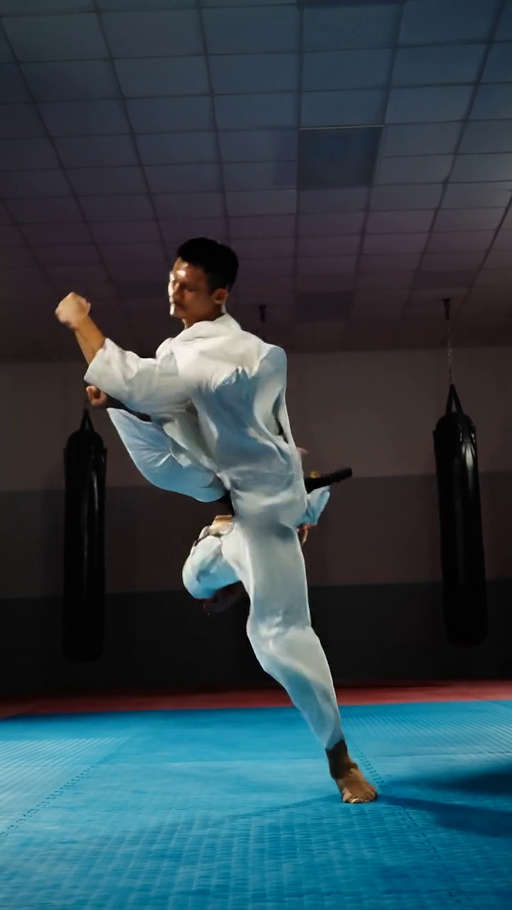} &
\begin{overpic}[width=\imgw,keepaspectratio]{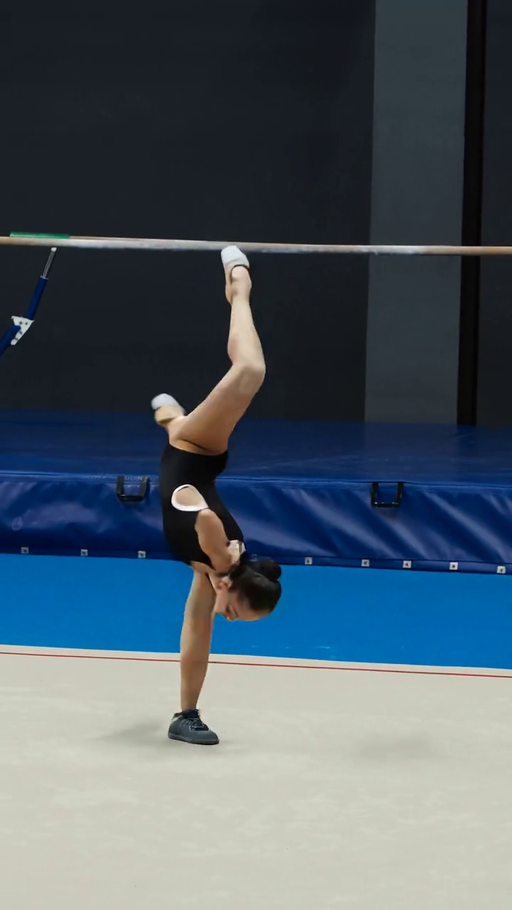}
  \put(3,5){\color{black}\tiny\bfseries Wan 2.1 14B}
\end{overpic} &
\includegraphics[width=\imgw,keepaspectratio]{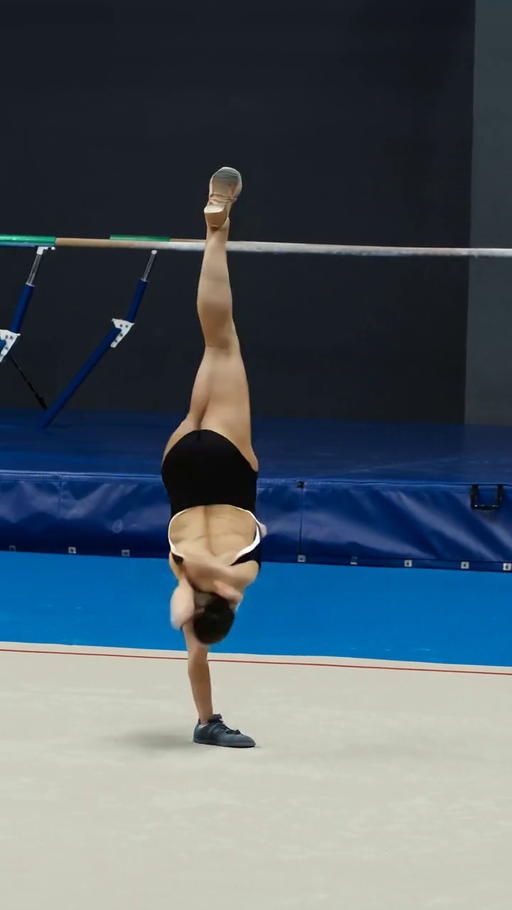} &
\includegraphics[width=\imgw,keepaspectratio]{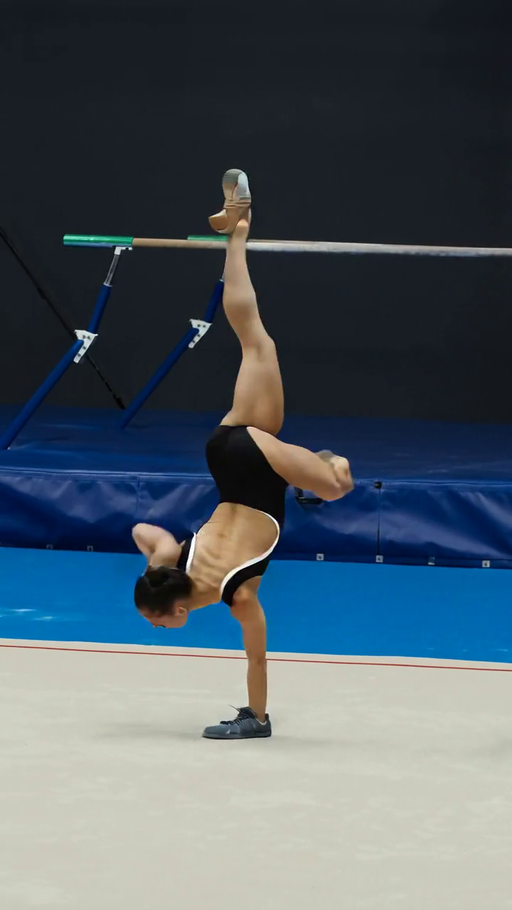}
\end{tabular}

\vspace{-0.2em}

\begin{tabular}{@{}*{3}{c}@{\hspace{\groupgap}}*{3}{c}@{\hspace{\groupgap}}*{3}{c}@{}}
\begin{overpic}[width=\imgw,keepaspectratio]{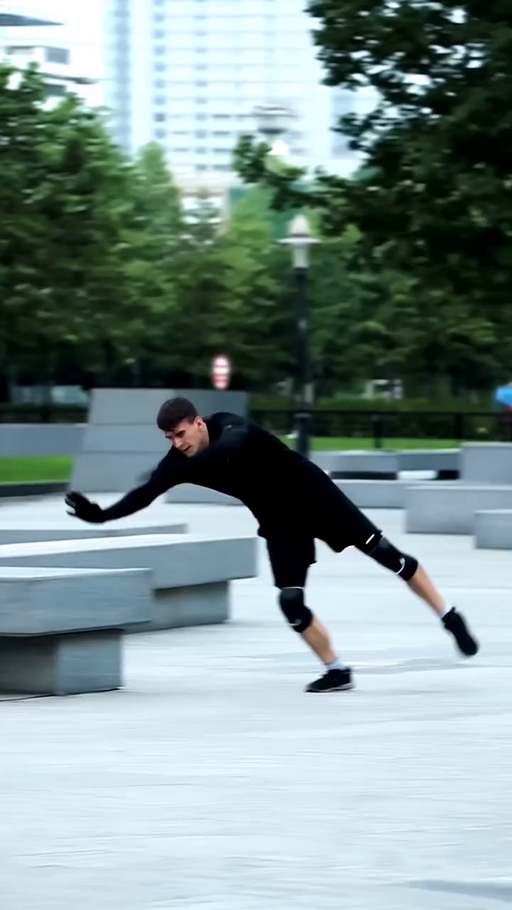}
  \put(3,5){\color{black}\tiny\bfseries Wan 2.1 14B + \OURS}
\end{overpic} & 
\includegraphics[width=\imgw,keepaspectratio]{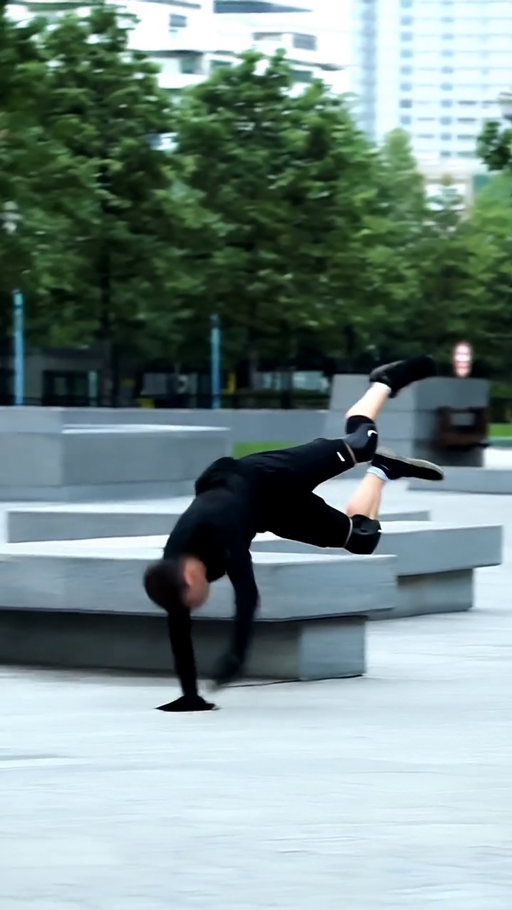} &
\includegraphics[width=\imgw,keepaspectratio]{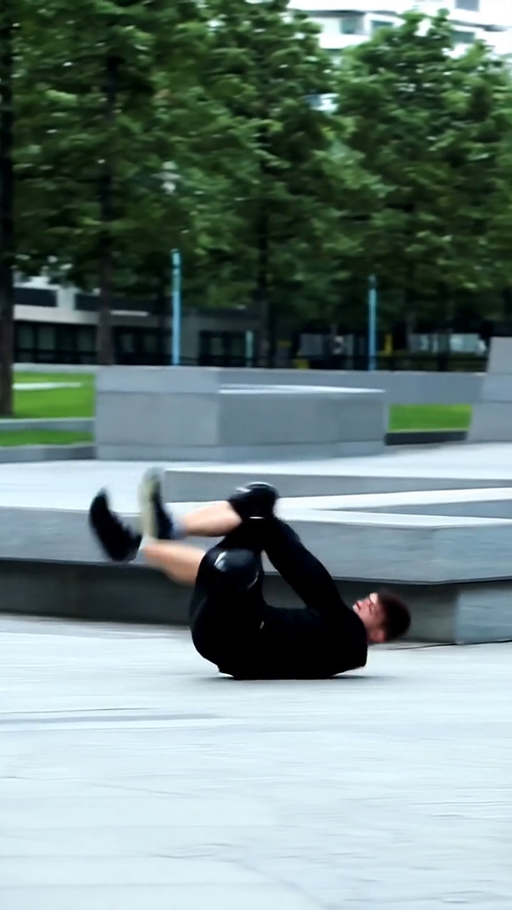} &
\begin{overpic}[width=\imgw,keepaspectratio]{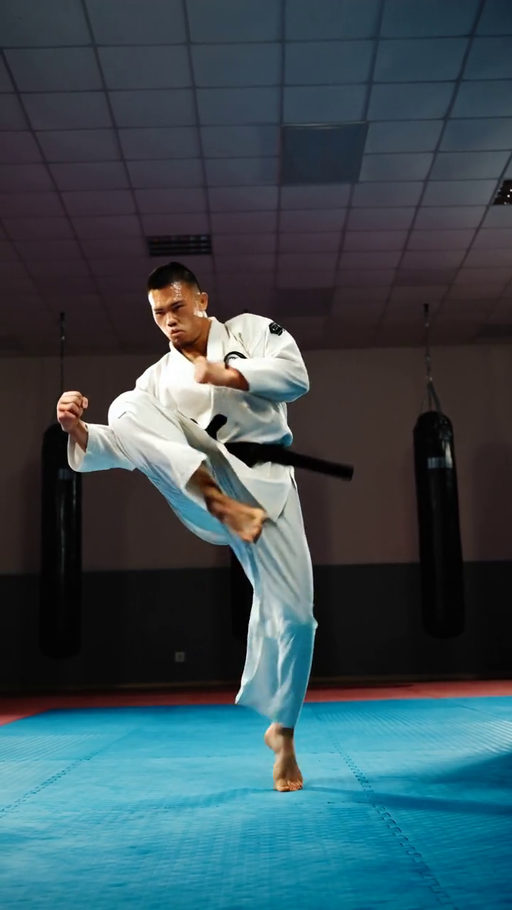}
  \put(3,5){\color{white}\tiny\bfseries Wan 2.1 14B + \OURS}
\end{overpic} & 
\includegraphics[width=\imgw,keepaspectratio]{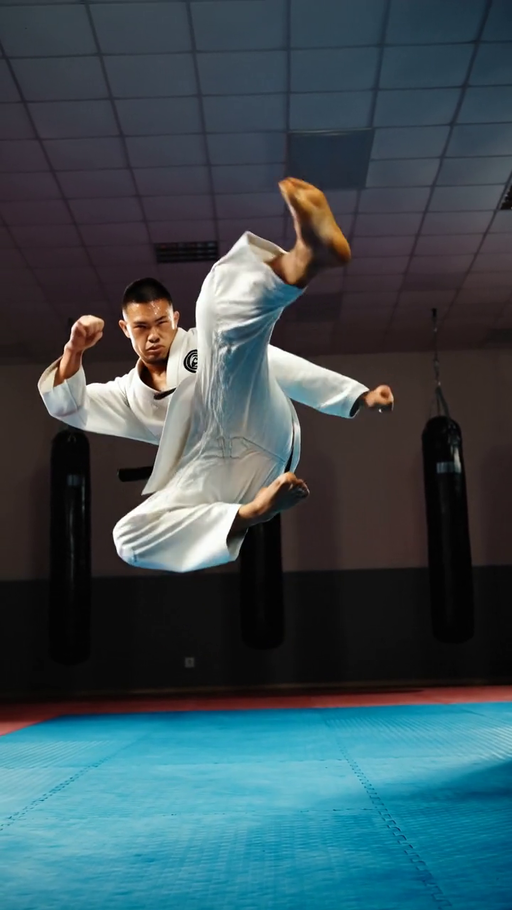} &
\includegraphics[width=\imgw,keepaspectratio]{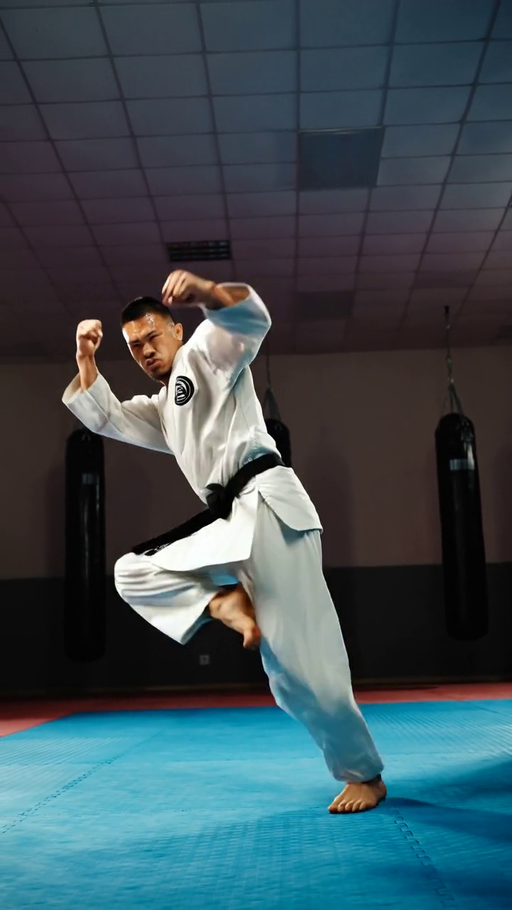} &
\begin{overpic}[width=\imgw,keepaspectratio]{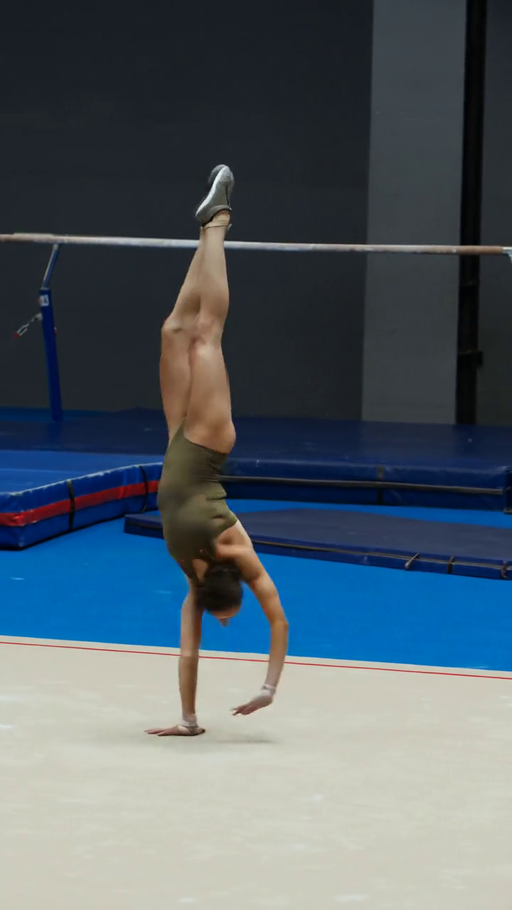}
  \put(3,5){\color{black}\tiny\bfseries Wan 2.1 14B + \OURS}
\end{overpic} & 
\includegraphics[width=\imgw,keepaspectratio]{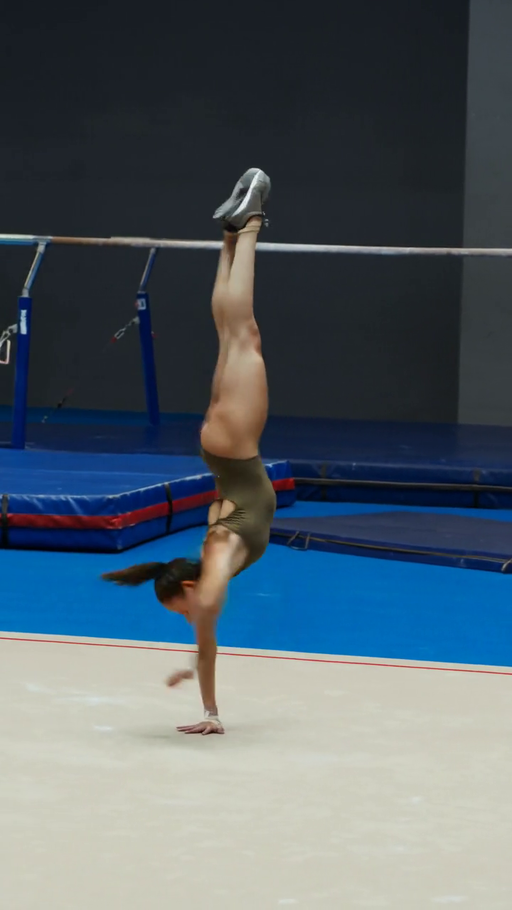} &
\includegraphics[width=\imgw,keepaspectratio]{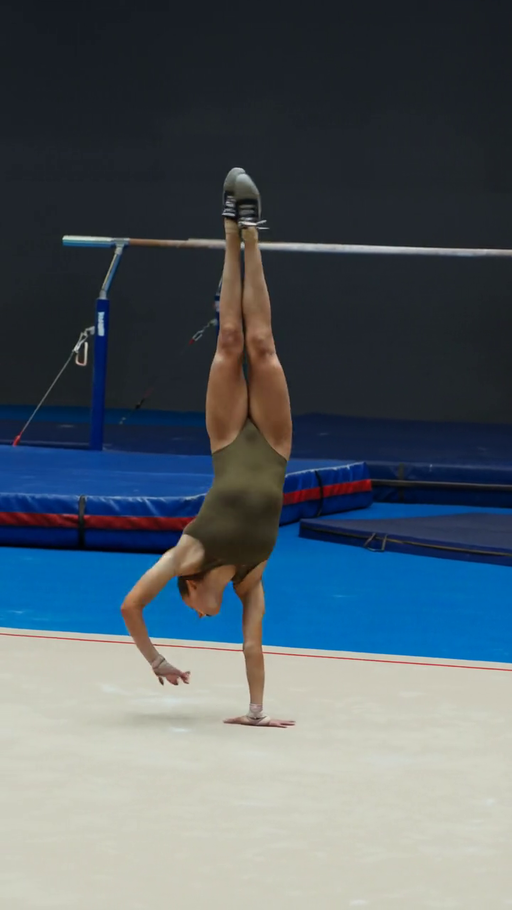}
\end{tabular}

\caption{\textbf{Qualitative comparison} of videos generated by our method (bottom) with Wan 2.1 14B~\cite{wan} (top). Wan 2.1 generates distorted poses (left), two-headed person (middle), and inconsistent legs (right), whereas Wan 2.1 trained with \OURS generates realistic humans.}
\label{fig:video-gen-results}
\vspace{-0.5cm}
\end{figure*}

Secondly, to establish that training with \OURS reward model improves video generation, we compare our outputs with baselines'. Similar to the above, we use the same human annotation setup. Additionally, we also ask the annotators if the generated video is faithful to the text prompt. This additional question is designed to ensure our training did not reduce the prompt faithfulness.
Now, we discuss the specifics of the two evaluations, followed by the results and ablations.

\subsection{Evaluating \OURS reward model}
\label{sec:reward-model-eval}

Recall that given two videos $\bm{V}_1$ and $\bm{V}_2$ where $\bm{V}_1$ shows better human appearance and motion, a good design of $R$ would imply $R(\bm{V}_1) > R(\bm{V}_2)$ for all $(\bm{V}_1, \bm{V}_2)$ pairs. Naturally, for a real video $\bm{V}$, $R(\bm{V})$ should be maximum. Overall, $R$ establishes a way to compare generated videos. Our protocol to evaluate \OURS is described in detail next.

\noindent \textbf{Generating videos with humans.} We use Wan 2.1~\cite{wan} 1.3B and 14B models, and generate $600$ video pairs, i.e., a total of $1200$ videos. We use the test set with $300$ motion descriptions, as discussed above.

\noindent \textbf{Evaluation metric.} As discussed above, we use human annotations to obtain ground truth preferences. We report accuracy as the fraction of samples where \OURS's relative order matches the preference of annotators' consensus.

\noindent \textbf{Results.} \cref{tab:winrate-expt1} shows the accuracy for all the methods. We outperform the best zero-shot method (VLM-as-a-judge) by more than $22\%$. We even outperform VBench-2.0~\cite{vbench2} by more than $3\%$, even though VBench-2.0 is trained with manually annotated data.
Finally, alternate visual/motion quality reward models are not suited for scoring human appearance in generated videos. In summary, our designed \OURS is strongly correlated with the human preferences. We use \OURS to train video diffusion model to generate better humans, as shown next.

\subsection{Video generation with \OURS}
\label{sec:video-gen-results}

In this section, we demonstrate utilizing our proposed \OURS to improve video generation. We use GRPO~\cite{deepseekmath,dancegrpo,flowgrpo}, with \OURS as the reward model to \emph{post-train} state-of-the-art video diffusion models.

\noindent \textbf{Training and testing prompt set.} We use the same training and testing prompt set, as described in the~\cref{sec:expt-setup}. Specifically for training Wan 2.1~\cite{wan} 1.3B, we choose medium and hard motions, whereas, we choose only hard motions for Wan 2.1~\cite{wan} 14B. This choice reflects the relative strength of the two variants since GRPO is more efficient with training samples having high variance.
Nevertheless, we evaluate on the whole testing set containing an equal split of easy, medium and difficult prompts.

\noindent \textbf{Evaluation metric.} In this experiment, we report win-rate between ours and the baseline as the fraction of times the annotator's preferred video was generated by \OURS.

\noindent \textbf{Results.} \cref{fig:qual} shows the results. Our method achieves a better win-rate compared to all the baselines, for all the prompt difficulty levels. Our average win-rate over the baselines is $73\%$ on the hard prompt set, $65\%$ for the medium prompt, and $63\%$ with the easy prompt set. These results showcase the strong improvement in human appearance and motion, using our zero-shot \OURS. Furthermore, the prompt faithfulness win-rate over the base Wan 2.1 is within $\pm 4\%$, implying that our training preserves the prompt faithfulness; detailed faithfulness result is in the~\cref{sec:appdx:addl_qual}.
Overall, we significantly improve human appearance, while preserving prompt faithfulness.

\noindent \textbf{Visualizations.} \cref{fig:video-gen-results} shows representative outputs from the GRPO training, and compares them with the base model (also see the~\cref{sec:appdx:videos} for video outputs). We see many anomalies in the base model that are absent in our generations. For example, a video of martial art 540-spinning kick shows two heads, whereas our generation correctly shows the action without any human appearance anomaly. We see the same improvement for all motion difficulty levels.

\subsection{Ablations}
\label{sec:ablations}

In this section, we present ablations to assess the contribution of each component and to show that our approach remains effective across varied design and training choices.
Each of the aspects are discussed next.

\begin{table}[t]\footnotesize
\centering
\begin{minipage}[t]{0.48\linewidth}
\vspace*{0pt}
\centering
\resizebox{\linewidth}{!}{
\begin{tabular}{lc}
\toprule
Reward model & Accuracy \\
\midrule
GroundingDINO~\cite{mmdetection} & 71.2 \\
Only P-score & 59.1 \\
Only H-score & 74.2 \\
Prompt-video sim. & 75.5 \\
Optical flow~\cite{dpflow} & 74.8 \\
Fewer frames (R) & 71.8 \\
Fewer frames (F) & 71.2 \\
\midrule
\OURS & \textbf{77.4} \\
\bottomrule
\end{tabular}}
\end{minipage}
\hfill
\begin{minipage}[t]{0.5\linewidth}
\vspace*{0pt}
\centering
\resizebox{\linewidth}{!}{
\begin{tabular}{lc}
\toprule
Reward model & \OURS Win-rate\\
\midrule
\OURS vs & \\
~~Only H-score & 75.0 \\
~~Prompt-video sim. & 57.0 \\
~~Optical flow~\cite{dpflow} & 90.0 \\
\bottomrule
\end{tabular}
}
\vfill
\resizebox{\linewidth}{!}{
\begin{tabular}{lc}
\toprule
Method & \OURS score\\
\midrule
Ours w/ KL-div & 0.961 \\
Ours w/o prompt exp. & 0.936 \\
\midrule
Ours & \textbf{0.975} \\
\bottomrule
\end{tabular}%
}
\end{minipage}
\caption{\textbf{Ablations}. On left, we compare \OURS to alternative reward functions described in~\cref{sec:reward-model-ablations} using accuracy measure from~\cref{sec:reward-model-eval}. On right (top), we compare generation performance with variants of \OURS as the reward model using win-rate. On right (bottom), we evaluate different training settings and the reward obtained on a validation set.
}
\label{tab:main-ablation}
\vspace{-0.5cm}
\end{table}

\subsubsection{\OURS ablations}\label{sec:reward-model-ablations}

We compare \OURS with the following:

\begin{itemize}
    \item Open-world object detector $\mathcal{D}$:  We replace the closed-vocabulary ViTDet~\cite{vitdet,detectron2} with an open vocabulary GroundingDINO~\cite{groundingdino,mmdetection} to evaluate if we can use class-independent open world object detector. We try various open world terms: `person', `human',`eyes', `legs', `hands', and a combination of these terms. We report the best performing combination.
    \item Only temporal prompt alignment score (Only P-score): In this ablation, we only use P-score as the reward model and do not use human detection score (H-score).
    \item Only human detection score (Only H-score): We investigate the role of P-score by removing it for this ablation.
    \item Prompt-video similarity as P-score: In this ablation, we replace our P-score with the average similarity score between the prompt and all the video frames. We use the same BLIP~\cite{blip} for similarity calculation.
    \item Optical flow as P-score: In this ablation, we explore if optical flow~\cite{dpflow} can serve as P-score, instead of a language-based similarity measure.
    \item Using fewer frames: In this ablation, we evaluate the reward on a subset of frames. Specifically, we explore two variants: randomly sampled ordered subset of $20\%$ frames (R), and a subset having the first $20\%$ frames (F).
\end{itemize}

\noindent \textbf{Results.} \cref{tab:main-ablation} (left) shows the results.
We report accuracy, as defined in the~\cref{sec:reward-model-eval}.
\OURS's accuracy outperforms all the other choices. In particular, despite the flexibility of open-world object detector, a closed vocabulary class-based object segmentation gives stronger performance. Next, using only P-score gives very low performance since action progress calculation is uncorrelated with human appearance quality. Conversely, if we only use  human detection score, the human appearance is very good but the motion is very conservative and often unfaithful to the input prompt. Next, prompt-video similarity can be high even with very little motion since overall prompt does not have sub-action descriptions. Moreover, optical flow also performs worse than \OURS due to the inability of optical flow to attend to human motion specifically. Lastly, using fewer frames degrades the performance significantly, verifying no single-frame bias and no over-reliance on the first few frames.

\begin{figure}[t]\footnotesize
    \centering
    \includegraphics[width=\linewidth]{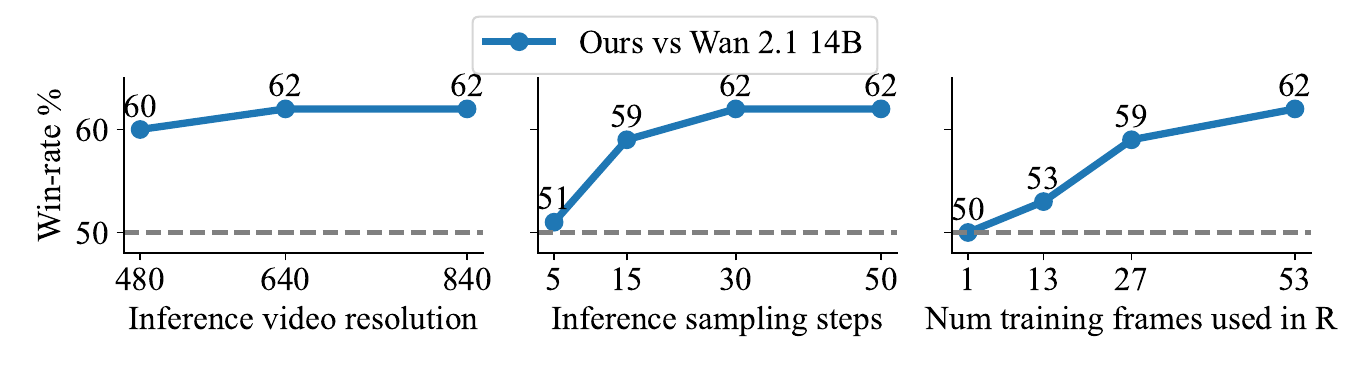}
    \caption{\textbf{Ablating design-choices in \OURS:}  inference video resolution (left), number of sampling steps (middle), and number of frames used to compute \OURS during training (right). We use human evaluation win-rate against the base video model. We find our gain is consistent across resolutions and increases marginally with inference sampling steps. Our performance significantly increases with more frames considered.}
    \label{fig:ablation-plot}
    \vspace{-0.5cm}
\end{figure}

\subsubsection{Ablations of video generation with \OURS}

We experiment with various choices in video generation post-training. We use the high-performing subset of reward models from the ablations above, and train GRPO using those reward models to further establish failures of those model choices.  All the experiments are done with Wan 2.1 14B on the easy prompt set. Specifically, we perform the following ablations, already described above: Only human detection score (only H-score), prompt-video similarity as P-score, and finally, optical flow as P-score. \cref{tab:main-ablation} (top right) shows the win-rate for the chosen ablations. We observe that \OURS is a better reward model than these ablations. Specifically, both human detection score (H-score) and temporal prompt alignment score (P-score) are crucial for improving humans in generated videos.

\newlength\groupgapB
\setlength\groupgapB{4pt} %

\newlength\imgv
\setlength\imgv{\dimexpr(\columnwidth-\groupgap)/4\relax}

\begin{figure}[t]\footnotesize
\centering
\setlength{\tabcolsep}{0pt}
\renewcommand{\arraystretch}{1}

\begin{tabular}{@{}*{4}{c}@{}}
\multicolumn{4}{@{}p{\dimexpr4\imgv\relax}@{}}{\centering\footnotesize \emph{A playful monkey jumping  on a grassy field.}} \\

\begin{overpic}[width=\imgv,keepaspectratio]{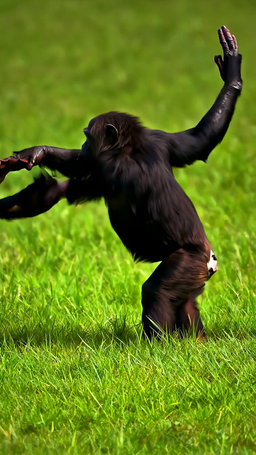}
  \put(3,5){\color{white}\scriptsize\bfseries Wan 2.1 14B}
\end{overpic} &
\includegraphics[width=\imgv,keepaspectratio]{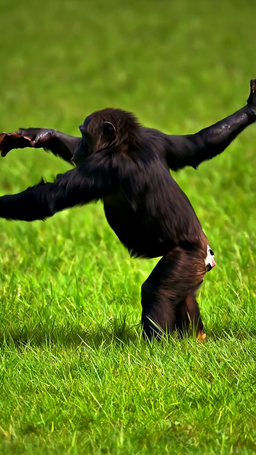} &
\includegraphics[width=\imgv,keepaspectratio]{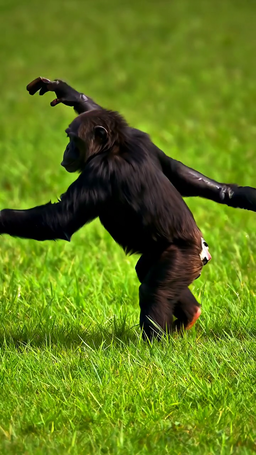} &
\includegraphics[width=\imgv,keepaspectratio]{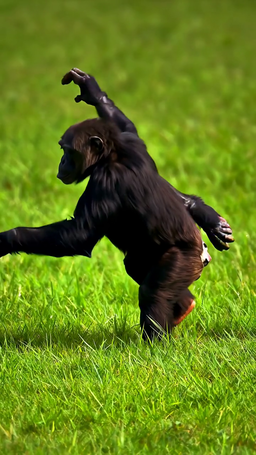}

\end{tabular}

\vspace{-0.2em}

\begin{tabular}{@{}*{4}{c}@{}}

\begin{overpic}[width=\imgv,keepaspectratio]{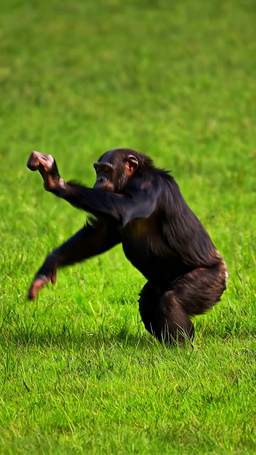}
  \put(0,5){\color{white}{\fontsize{6pt}{7pt}\selectfont\bfseries Wan 2.1 14B + \OURS}}
\end{overpic} &
\includegraphics[width=\imgv,keepaspectratio]{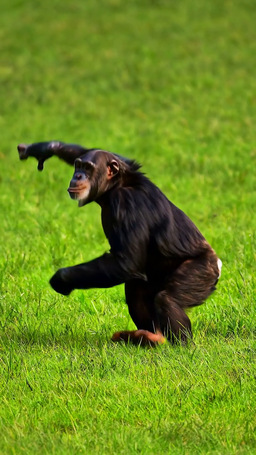} &
\includegraphics[width=\imgv,keepaspectratio]{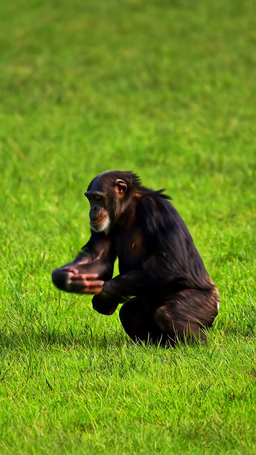} &
\includegraphics[width=\imgv,keepaspectratio]{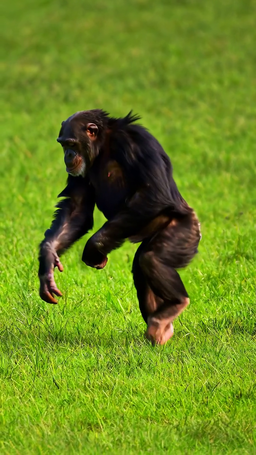}

\end{tabular}

\caption{\textbf{Model trained with \OURS generates better zero-shot animal videos,} 
whereas the base model generates videos with extra/missing body parts and unrealistic poses, showing that \OURS enables the model to generate realistic body deformations.}
\label{fig:animals}
\vspace{-0.5cm}
\end{figure}

Next, we ablate our choices in the GRPO training setup. We investigate the affect of using KL-divergence in the GRPO training, and that of not using prompt expansion. \cref{tab:main-ablation} (bottom right) shows the performance. We simply use the reward score, \ie, \OURS to evaluate the performance since the reward model is the same in all the methods being compared.
We observe that KL-divergence is not very helpful in the performance, while requiring more memory since we need to maintain a copy of the original policy. Moreover, we do not see an improvement in GRPO training without prompt expansion.

Lastly, we present our results with various video parameters to showcase that our performance gain is agnostic to those model choices:

\begin{itemize}
    \item Impact of inference video resolution: During inference, we generate videos with different resolutions to show that the improved performance is independent of the resolution. We consider an aspect ratio of $1$.
    \item Impact of the number of inference steps: We generate videos with varying numbers of steps to show that the performance gain is consistent across different step counts.
    \item Impact of the number of training frames considered: We investigate if all frames are required when calculating \OURS. In place of $53$ training frames, we consider $1, 13$, or $27$ frames during training.
\end{itemize}

\cref{fig:ablation-plot} shows ablation plots for various training and inference choices. We observe that our gain is significant for all video sizes, for all reasonable number of sampling steps. Finally, an increasing win-rate when more frames are considered signals no single-frame bias in \OURS.

\subsection{Beyond human video generation}

While \OURS is designed for optimizing human video generation, we find the benefits
with \OURS extend beyond that. In particular, we find that a model trained with \OURS even improves generation of animals, and human-object interactions. We discuss these next.

\noindent \textbf{Generating videos with animals.} We observe that the deformations in humans are similar to those in animals such as dogs, cats, and monkeys. Although \OURS is not specifically designed for animal detection, videos generated by our trained model exhibit the correct animal motions.

We use Llama-3.2 8B~\cite{llama3} to create a prompt set for this experiment. We use dog, cat, chimpanzee, monkey, ape, gorilla, and orangutan as the subject and prompt the language model to generate $100$ prompts that shows various actions like backflip, jump, cartwheel, etc. Specifically, actions that are not typical characteristics of these animals are also included to test the generation ability.
\cref{fig:animals} illustrates one such case. The base model outputs a monkey with three hands, while our method produces the correct anatomy. More examples in the~\cref{sec:appdx:videos}.

\newlength\groupgapA
\setlength\groupgapA{4pt} %

\newlength\imgz
\setlength\imgz{\dimexpr(\columnwidth)/5\relax}

\begin{figure}[t]\footnotesize
\centering
\setlength{\tabcolsep}{0pt}
\renewcommand{\arraystretch}{1}

\begin{tabular}{@{}c@{\hspace{\groupgapA}\hspace{\groupgapA}}*{4}{c}@{}}
Detector $\mathcal{D}$ & \multicolumn{4}{@{}p{\dimexpr4\imgz\relax}@{}}{\centering\footnotesize \emph{A professional tennis player executing a powerful swing.}} \\

\begin{overpic}[width=\imgz,keepaspectratio]{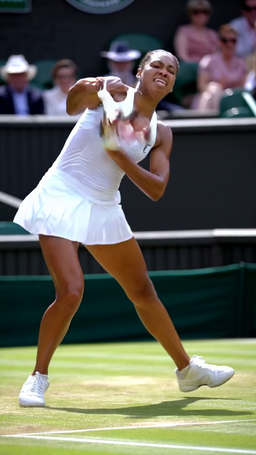}
    \put(3,10){\color{red}\framebox(48,80){}}
  \put(20,3){\color{red}\tiny\bfseries 0.92}
\end{overpic} &

\begin{overpic}[width=\imgz,keepaspectratio]{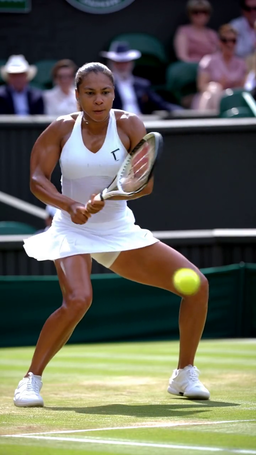}
  \put(3,3){\color{black}\tiny\bfseries Wan 2.1 14B}
\end{overpic} & 
\includegraphics[width=\imgz,keepaspectratio]{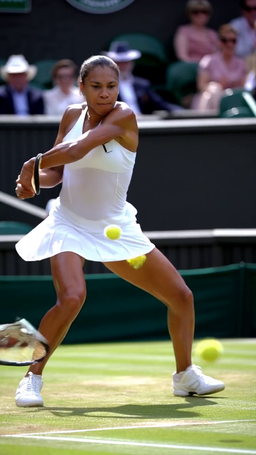} &
\includegraphics[width=\imgz,keepaspectratio]{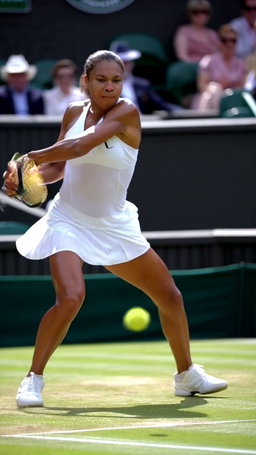} &
\includegraphics[width=\imgz,keepaspectratio]{figs/output_videos/OI0-0/frame_00078_256x.png} 

\end{tabular}

\vspace{-0.2em}

\begin{tabular}{@{}c@{\hspace{\groupgapA}\hspace{\groupgapA}}*{4}{c}@{}}

\begin{overpic}[width=\imgz,keepaspectratio]{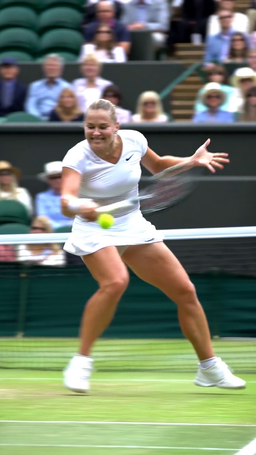}
    \put(12,10){\color{blue}\framebox(42,70){}}
  \put(20,3){\color{blue}\tiny\bfseries 1.0}
\end{overpic} & 

\begin{overpic}[width=\imgz,keepaspectratio]{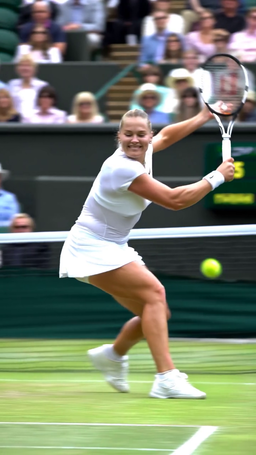}
  \put(0,3){\color{black}\tiny\bfseries Wan 2.1 14B + \OURS}
\end{overpic} & 
\includegraphics[width=\imgz,keepaspectratio]{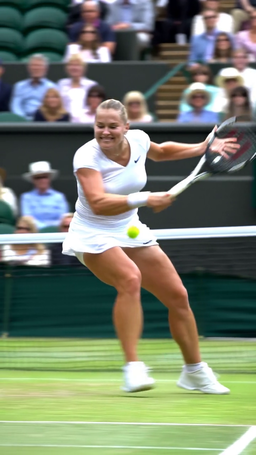} &
\includegraphics[width=\imgz,keepaspectratio]{figs/output_videos/OI0-1/frame_00044_256x.png} &
\includegraphics[width=\imgz,keepaspectratio]{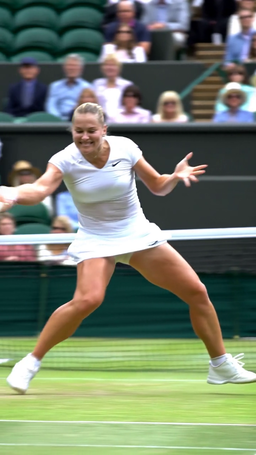}

\end{tabular}

\caption{\textbf{Example of improvement in human-object interaction.} Reward from our \OURS is higher when human-object interactions are correct (left). Training video generation with \OURS improves human-object interaction in generated videos (right).}
\label{fig:hoi}
\vspace{-0.5cm}
\end{figure}

\noindent \textbf{Improved human-object interaction.} Improving human appearance also improves human-object interactions since the two are tied together in natural videos. We observe that the human detection score from $\mathcal{D}$ is low in videos with poor human-object interactions, because of worse human hand and/or body appearance. For example, in \cref{fig:hoi} (left), we see a lower detection score when the tennis racket and hand is distorted, as opposed to when they are correctly generated. Consequently, we find that our model trained with \OURS has a nice side-effect of even generating better human-object interactions compared to the baseline model. \cref{fig:hoi}
(right) shows one such example in tennis, more examples in the~\cref{sec:appdx:videos}.
In summary, the advantages of \OURS goes beyond improvement in human appearance.

\section{Conclusion}
\label{sec:conclusion}

In this work, we propose \OURS, a remarkably simple reward model that can correctly quantify the human appearance, motion, and action progress. We first show that this reward model correlates well with human preference. Our zero-shot method's gain far exceeds other zero-shot methods, and even outperforms models trained with manually annotated data. Next, we use the proposed reward model \OURS to improve video generation with GRPO. As observed from the human preference studies, videos generated by our method is preferred by human raters over all other competing baselines. Lastly, we show that our training helps the model better understand deformable objects, thus generalizing the improvements to human-object interactions and animal video generation.

{
    \small
    \bibliographystyle{ieeenat_fullname}
    \bibliography{main}
}
\appendix
\clearpage
\setcounter{page}{1}
\maketitlesupplementary

\section{Videos for all figures in the main paper}\label[appx]{sec:appdx:videos}

We attach an HTML webpage `videos.html' that contains qualitative results, including videos for the figures in the main paper.

\section{Human evaluation Setup}\label[appx]{sec:appdx:human_eval}

\begin{figure}[t]\footnotesize
    \centering
    \includegraphics[width=\columnwidth]{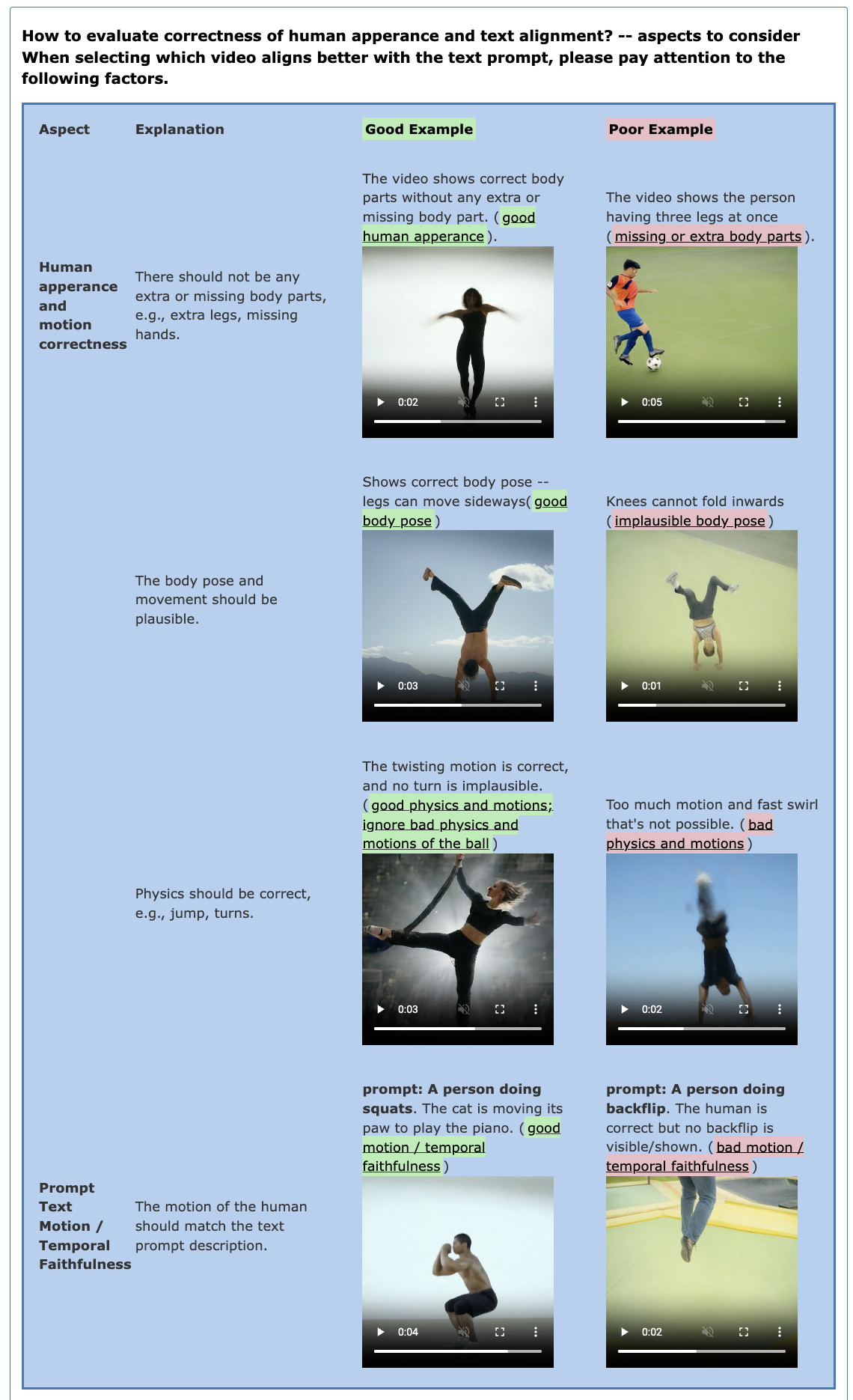}
    \caption{\textbf{Annotation interface instructions.}
    We describe the various axes of judging human appearance and motion, and provide good and bad examples for each of them.
    }
    \label{fig:amt_instructions}
    \vspace{-0.5cm}
\end{figure}

\cref{fig:amt_instructions} shows the instructions provided to the human annotators. We first show good and bad examples of various aspects of human appearance and motion. We focus on extra or missing body parts, pose sequence and movements, and physics. Next, we show an example of good and bad prompt faithfulness. These examples are visible to the annotators at all times, for their quick reference.

\begin{figure}[t]\footnotesize
    \centering
    \includegraphics[width=\columnwidth]{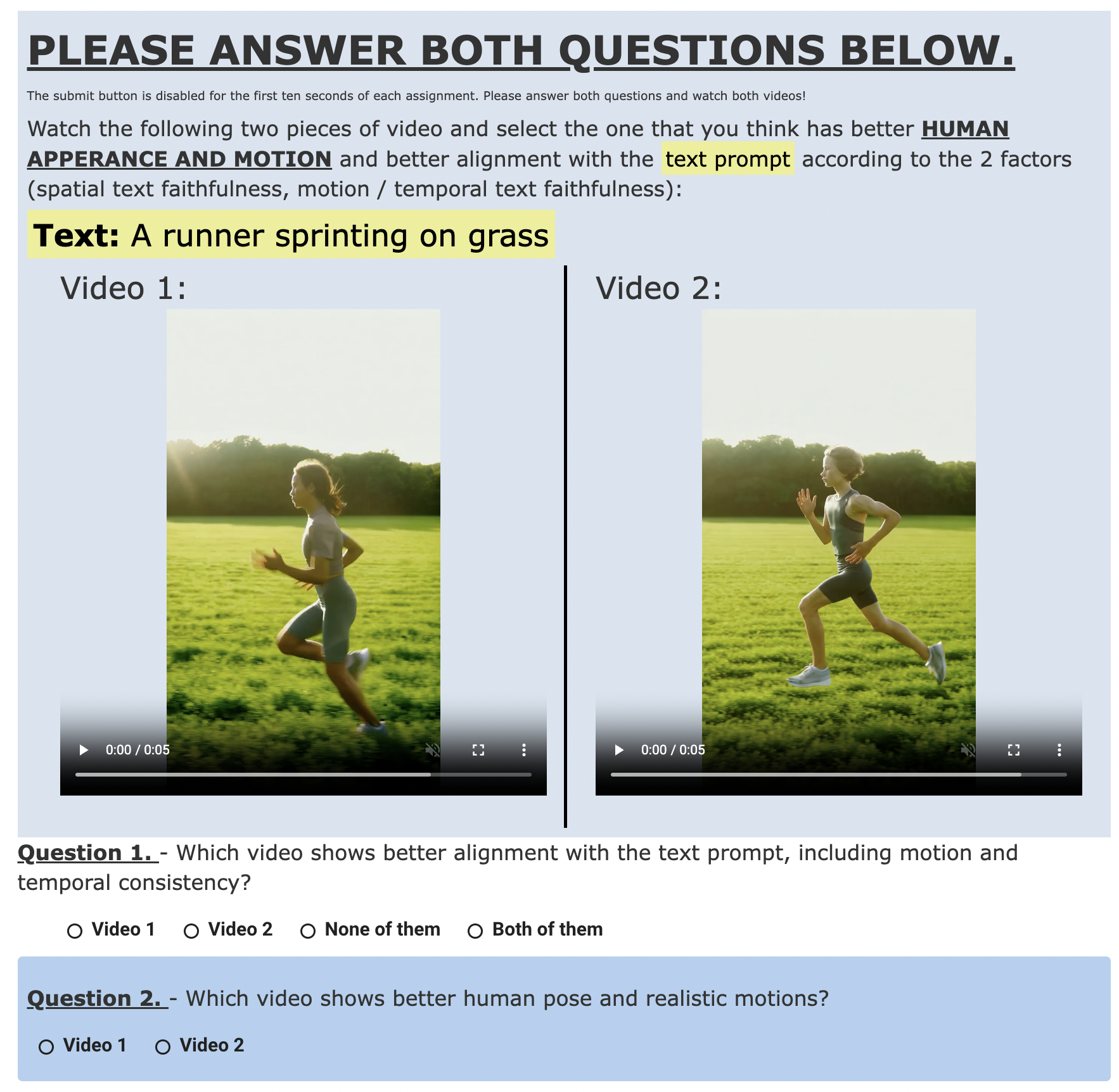}
    \caption{\textbf{Annotation questions.}
    We provide the two videos being compared, and the text prompt, and ask the annotators two questions. The first question asks if one, both, or none of the videos are faithful to the input prompt. The second question asks which of the two videos shows more realistic human appearance.
    }
    \label{fig:amt_questions}
    \vspace{-0.5cm}
\end{figure}

Next, \cref{fig:amt_questions} shows the annotation questions. We first ask whether one, both, or neither of the videos are faithful to the prompt. This question helps us assess whether the generated videos are meaningful. We do not want improvements in human appearance to come at the expense of prompt faithfulness. The second question asks which video has the better human appearance. Unlike the first question, we do not include a neutral option, since even when the differences are minimal, we want annotators to choose the video with better human appearance.

\section{LLM prompts}\label[appx]{sec:appdx:llm_prompts}

The following snippet contains the LLM prompt to generate $3000$ human motion sequence labels, as discussed in~\cref{sec:expt-setup}. To ensure the response length does not exceed the maximum new tokens limit, we ask the LLM to generate motion sequences for each difficulty level separately, \ie, $1000$ output motions per response.

\begin{tcolorbox}[breakable, boxrule=0mm]
Generate a list of 3000 human movement concepts and categorize them into three difficulty levels based on how physically challenging, dynamic, and coordinated the movement is.

Use the following structure:\\
- easy (1000 items): simple motions like sitting, walking, waving\\
- medium (1000 items): moderately dynamic motions like running, jumping, crawling, kicking\\
- hard (1000 items): acrobatic, athletic, or highly complex movements like flips, vaults, breakdancing moves\\

Requirements:\\
- Focus ONLY on human actions that require a full-body view (no micro-actions like winking, blinking, finger tapping, eye movement, typing fingers, etc.).\\
- Return the result as valid JSON, exactly in the format:\\

\{\\
  ``easy": [``concept1", ...],\\
  ``medium": [``concept1", ...],\\
  ``hard": [``concept1", ...]\\
\}\\

- Avoid duplicates strictly across all categories.\\
- Ensure the list is highly diverse across daily actions, sports, athletics, dance, martial arts, gymnastics, acrobatics, parkour, yoga, pilates, diving, circus arts, rehabilitation movements, etc.\\
- Concepts should be names only (no descriptions).\\
- Generate the sets one category at a time.\\
- Start now by generating the 1000 concepts for the hard category only.
\end{tcolorbox}

Next, the following snippet contains the LLM prompt to generate action phases $T_i$ given a text prompt, as discussed in~\cref{sec:method}:

\begin{tcolorbox}[breakable, boxrule=0mm]

You are an expert video coach and scriptwriter. Given a high-level caption for a video-generation prompt, break it into exactly 5 temporally ordered, visually concrete micro-phases that together span ~5 seconds total around the core action (NOT much longer than that).\\

STRICT TEMPORAL SCOPE:\\
- Model exactly a ~5 second window centered on the core action.\\
- Each phase should represent ~0.5-1.5 seconds (max 2 seconds).\\
- Do NOT include pre-roll/setup (e.g., “walks into the court”, “approaches the plate”) or post-roll (e.g., “leaves the scene”).\\
- Start at the moment the action preparation begins (e.g., split-step or load) and end right after follow-through/recovery.\\

VISUAL CONCRETENESS:\\
- Phases must be camera-visible body/object states and motions (stance, weight shift, unit turn, racquet/arm path, contact, follow-through), not narrative fluff.\\
- 1-2 sentences per phase, present tense, specific nouns/verbs, no questions.\\
- Maintain tight temporal continuity and causal flow across phases.\\
- If the caption is ambiguous, choose a plausible concrete interpretation centered on the core action in 5 seconds.\\

FORMAT (JSON only, no markdown or extra text):\\
\{ \\
  ``caption": ``\{original caption\}", \\
  ``phases": [ \\ 
    {``id": 1, ``name": ``\{short micro-phase label\}", ``description": ``$<$ 1-2 sentences, $\leq 2s$ of action>"}, \\
    {``id": 2, ``name": ``...", ``description": ``..."}, \\
    {``id": 3, ``name": ``...", ``description": ``..."}, \\
    {``id": 4, ``name": ``...", ``description": ``..."}, \\
    {``id": 5, ``name": ``...", ``description": ``..."} \\
  ] \\
\} \\

NEGATIVE EXAMPLES (DO NOT DO): \\
- “The player walks into the court…” (too early; out of 5s window) \\
- “He celebrates and exits the frame.” (post-roll; out of 5s window) \\
- Vague phrases like “then he moves” without body mechanics. \\

QUALITY TARGET (for something like “Rafael Nadal plays a forehand”): \\
- Phase 1 could be split-step + unit turn/load, \\
- Phase 2 racquet take-back + hip/shoulder coil, \\
- Phase 3 step/weight transfer + swing initiation, \\
- Phase 4 contact + extension, \\
- Phase 5 follow-through + balanced recovery.

\end{tcolorbox}

Finally, the following snippet contains the LLM prompt for the VLM-as-a-judge baseline, as introduced in~\cref{sec:expt-setup}. We ask $4$ questions, and take the score as the worst `no' logit probability, since the human appearance is as good as the worst aspect.

\begin{tcolorbox}[breakable, boxrule=0mm]

The video lasts for ``\{time of the video\}'' seconds, and ``\{number of frames\}'' frames are uniformly sampled from it. These frames are located at ``\{frame timesteps\}''. Return yes or no only for the following questions: \\ 

Q1. Across the entire video, does the person's pose sequence remain realistic without unnatural twists or left-right flips?\\
Q2. Are all body parts present or absent in ways that look natural, with no unnatural extra or missing parts beyond normal occlusion or clipping?\\
Q3. Do all body parts remain clearly distinct from the background throughout the video, without dissolving or fading away, even if they clip or go out of frame?\\
Q4. Does the video show realistic physics and natural human-object interactions?

\end{tcolorbox}

\section{Additional Results and Ablations}\label[appx]{sec:appdx:addl_qual}

\begin{table}[t]\footnotesize
\centering
\begin{tabular}{l c}
\hline
\textbf{Method} & \textbf{Accuracy} \\
\hline
VLM-as-a-judge~\cite{qwen2.5}\quad & 55.0 \\
DanceGRPO~\cite{dancegrpo} & 51.5 \\
\midrule
VBench-2.0~\cite{vbench2} & 72.7 \\
\rowcolor{Gray}
2D pose~\cite{vitpose} & 62.3 \\
\rowcolor{Gray}
3D pose~\cite{d3dp} & 63.4 \\
\rowcolor{Gray}
Optical flow~\cite{dpflow} & 60.2 \\
\midrule
\OURS (Ours) & \textbf{77.4} \\
\hline
\end{tabular}
\caption{\textbf{Accuracy} of additional reward models (gray rows) in predicting human-preferred videos. Our \OURS, using human detection confidence, outperforms models using alternate modalities, \ie, 2D/3D pose, and optical flow.}
\label{tab:winrate-expt1-supp}
\vspace{-0.5cm}
\end{table}

\subsection{Additional \OURS comparisons}

\begin{figure}[t]\footnotesize
    \centering
    \includegraphics[width=\columnwidth]{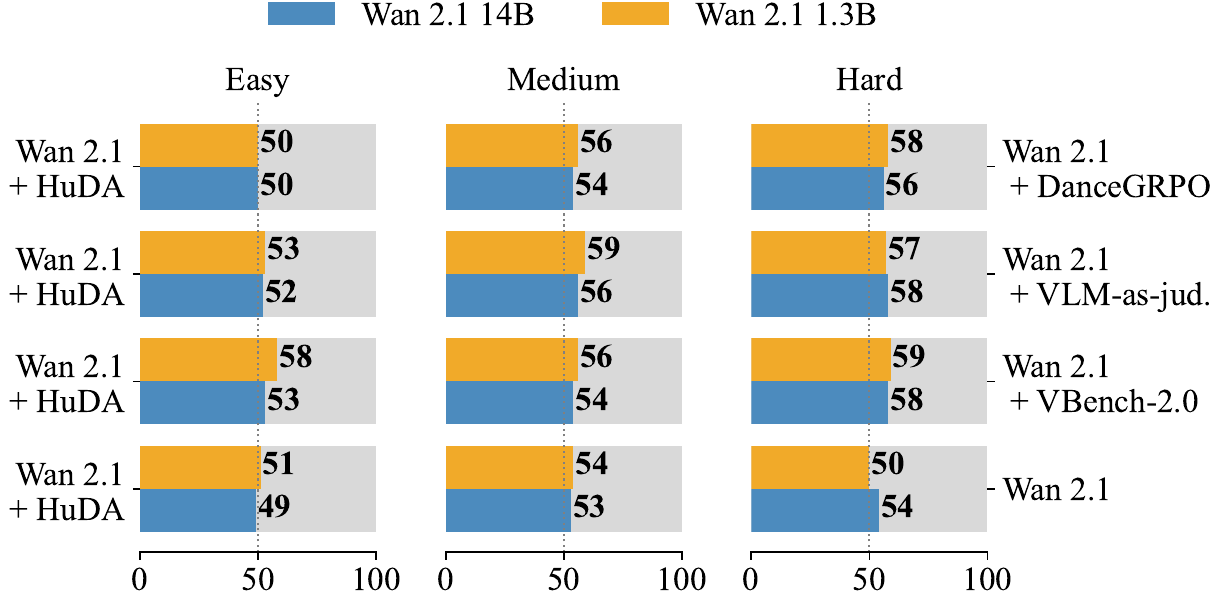}
    \caption{\textbf{Prompt faithfulness win-rates.}
    Prompt faithfulness comparison between baselines and our model trained with \OURS as the reward model. Despite \OURS’s focus on improving human appearance, the prompt faithfulness win-rate remains within $47\text{–}54\%$, showing that prompt faithfulness is preserved. In particular, the temporal prompt alignment module (\cref{sec:method}) in \OURS helps maintain faithfulness.
    }
    \label{fig:qual_faithfulness}
    \vspace{-0.5cm}
\end{figure}

In this section, we show results with additional reward models, as part of our early explorations in this direction. We explore the following reward models:

\begin{itemize}
    \item 2D pose~\cite{vitpose}: In this baseline, we employ a video to 2D pose method~\cite{vitpose} in inference mode, and obtain 2D pose sequences for real and generated videos. The real videos are obtained from the COCO~\cite{coco} dataset, and the generated videos are obtained from our prompt set, with Wan 2.1~\cite{wan} 1.3B and 14B. The output poses are used to train a transformer model to output a class label, with CLS token used as the aggregate representation.
    \item 3D pose~\cite{d3dp}: This baseline is similar to the above 2D pose baseline, except we use inferred 3D poses as input to the transformer architecture.
    \item Optical flow~\cite{dpflow}: We extract optical flow from real and generated videos. The videos are the same as discussed in the 2D pose baseline.
\end{itemize}

\cref{tab:winrate-expt1-supp} shows the results. We see that attempting this task in an alternate modality, \ie, pose or flow, is not beneficial. The key issue is that the transformation from the pixel space to the pose or the flow space assumes correct human appearance, since all the models trained for these tasks use only real videos. Therefore, unlike pixel-space, where we see the effectiveness of human detector~\cite{vitdet,detectron2} in zero-shot setting, pose and flow modalities are not appropriate reward models for this task.

\subsection{Faithfulness of videos generated with \OURS}

\newlength\groupgapF
\setlength\groupgapF{4pt} %

\newlength\imgf
\setlength\imgf{\dimexpr(\columnwidth)/2\relax}

\begin{figure}[t]\footnotesize
\centering
\setlength{\tabcolsep}{0pt}
\renewcommand{\arraystretch}{1}

\begin{tabular}{@{}*{2}{c}@{}}
\multicolumn{2}{@{}p{\dimexpr2\imgf\relax}@{}}{\centering\footnotesize \emph{Detector $\mathcal{D}$ scores.}} \\

\begin{overpic}[width=\imgf,keepaspectratio]{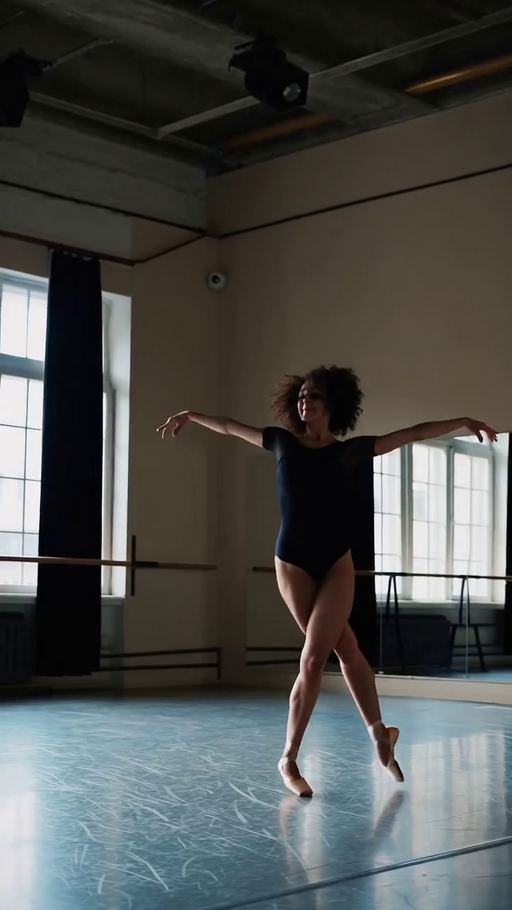}
    \put(16,11){\color{blue}\framebox(39,50){}}
  \put(34,7){\color{blue}\small\bfseries 1.0}
\end{overpic} & 

\begin{overpic}[width=\imgf,keepaspectratio]{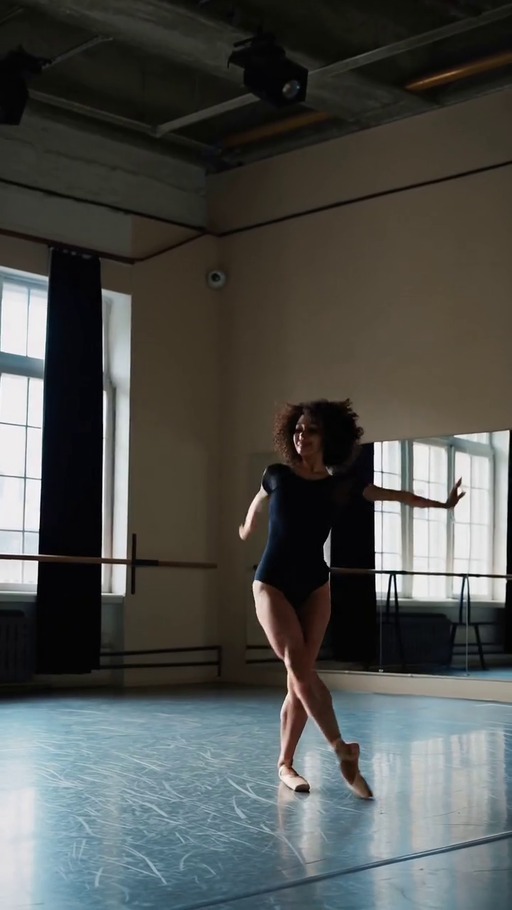}
    \put(23,11){\color{blue}\framebox(28,47){}}
  \put(37,7){\color{blue}\small\bfseries 1.0}
\end{overpic}

\end{tabular}

\vspace{-0.2em}

\caption{\textbf{Capturing fast changes of direction is challenging.} The crossing of legs changes rapidly within a few frames, indicating implausible motion. Both frames show correct pose, and hence, \OURS does not capture this motion implausibility. We observe that all reward models fail for this case.}
\label{fig:failure}
\vspace{-0.5cm}
\end{figure}

In addition to ensuring an improvement in human appearance, we compare the prompt faithfulness between videos generated with \OURS and the baselines. Comparing faithfulness is crucial since any improvement in human appearance is not helpful if the prompt faithfulness decreases. For example, generating a static person will always give a high human appearance score. However, such a scenario is undesirable. At the same time, our focus is on improving human appearance. Therefore, we want the prompt faithfulness to be as good as the model without training with \OURS, while significantly increasing the human appearance win-rate. 

\cref{fig:qual_faithfulness} shows the results. We observe that compared to Wan 2.1~\cite{wan}, the same model trained with \OURS has a prompt faithfulness win-rate between $48\%$ to $54\%$ across the three testing splits (easy, medium, and hard). Other reward models, \ie, Wan 2.1 + VLM-as-a-judge and VBench-2.0~\cite{vbench2} do not use temporal prompt alignment score, and hence, our prompt faithfulness win-rate with them is higher.

\section{Limitation}

While \OURS improves human appearance across diverse settings, we observe some instances where the output score from \OURS is high for videos with imperfect human appearance. In particular, since human detection score aggregates frame-level scores, human appearance inconsistency during fast movements is not captured by \OURS. \cref{fig:failure} shows one such example. The crossing of legs changes within a few frames, indicating implausibility. Each individual frame by itself is correct. Hence, \OURS incorrectly scores this video high. The same limitation is observed in all the reward models we compare against. However, we observe very few cases of this type of implausibility.

\end{document}